%% file: acl2023.tex
\pdfoutput=1
\documentclass[11pt]{article}

\usepackage[]{ACL2023}

\usepackage{times}
\usepackage{latexsym}

\usepackage[T1]{fontenc}
\usepackage[utf8]{inputenc}

\usepackage{microtype}

\usepackage{inconsolata}

\usepackage{adjustbox}
\usepackage{graphicx}
\usepackage{amssymb,amsmath,array}
\usepackage{pifont}
\usepackage{multirow,makecell}
\usepackage{url}
\usepackage{latexsym}
\usepackage{array}
\usepackage{todonotes}
\usepackage{subfigure} 
\usepackage{algorithm}
\usepackage{algorithmic}
\usepackage{makecell}
\usepackage{booktabs}
\usepackage{listings}
\usepackage{csquotes}
\usepackage{amsthm}
\usepackage{bm}
\usepackage{xcolor,colortbl}
\usepackage{xspace}
\usepackage{cleveref}
\crefformat{section}{\S#2#1#3}
\crefformat{subsection}{\S#2#1#3}
\newcommand\sect[1]{\S\ref{#1}}

\newcommand\imdb{\textsc{IMDB}\xspace}
\newcommand\ag{\textsc{AG News}\xspace}
\newcommand\amazon{\textsc{Amazon Review}\xspace}

\newcommand\sst{\textsc{SST-2}\xspace}
\newcommand\yahoo{\textsc{Yahoo! Answer}\xspace}

\newcommand\vat{\textsc{VAT}\xspace}
\newcommand\uda{\textsc{Uda}\xspace}
\newcommand\stl{\textsc{Stl}\xspace}
\newcommand\fixmatch{\textsc{FixMatch}\xspace}
\newcommand\dash{\textsc{Dash}\xspace}
\newcommand\flexmatch{\textsc{FlexMatch}\xspace}
\newcommand\adamatch{\textsc{AdaMatch}\xspace}
\newcommand\tapt{\textsc{Tapt}\xspace}
\newcommand\pt{\textsc{Pt}\xspace}
\newcommand\ft{\textsc{Ft}\xspace}
\newcommand\st{\textsc{St}\xspace}
\newcommand\ssl{\textsc{Ssl}\xspace}
\newcommand\nlp{\textsc{NLP}\xspace}
\newcommand\mlm{\textsc{Mlm}\xspace}
\newcommand\partials{\textsc{Supervised}\xspace}
\newcommand\fullys{\textsc{Fully-Supervised}\xspace}
\newcommand\roberta{\textsc{RoBERTa}\xspace}
\newcommand\robertabase{\textsc{RoBERTa-}{\scriptsize \textsc{Base}}\xspace}

\definecolor{cid}{HTML}{dae8f5}
\definecolor{ccon}{HTML}{fee9d4}
\definecolor{gred}{HTML}{cc0200}
\definecolor{ggreen}{HTML}{4C9F26}

\newcommand\hl{\cellcolor{cid}}
\newcommand\se{\cellcolor{ccon}}

\newcommand{\up}[1]{\tiny~(\textcolor{ggreen}{$\blacktriangle$}#1$\%$)}
\newcommand{\down}[1]{\tiny~(\textcolor{gred}{$\blacktriangledown$}#1$\%$)}
\title{Rethinking Semi-supervised Learning with Language Models}
\author{Zhengxiang Shi$^{^1}$\thanks{~~This work was done during an internship at Amazon, Alexa Shopping.} ~ Francesco Tonolini$^2$ ~ Nikolaos Aletras$^{2,3}$ ~ Emine Yilmaz$^{1,2}$ \\ {\bf Gabriella Kazai}$^2$ ~ {\bf Yunlong Jiao}$^2$\\
        $^1$ University College London, London, United Kingdom \\ 
        $^2$ Amazon, London, United Kingdom \\ 
        $^3$ University of Sheffield, Sheffield, United Kingdom \\
        \texttt{zhengxiang.shi.19@ucl.ac.uk} \\
        \texttt{\{tonolini,eminey,aletras,gkazai,jyunlong\}@amazon.com}
}
\begin{document}
\maketitle

\begin{abstract}
\input{paper/1_abstract.tex}
\end{abstract}

\section{Introduction}
\input{paper/2_introduction.tex}

\section{Preliminaries}
\input{paper/3_background}

\section{Experimental Setup}
\label{sec:setup}
\input{paper/4_setup}

\section{\st vs \tapt}
\label{sec:st_vs_tapt}

\input{paper/5_st_vs_tapt}

\section{Exploring the limits of \st and \tapt}
\label{sec:explore_limits}
\input{paper/6_analysis.tex}

\section{Domain Adaptation}
\label{sec:da}
\input{paper/7_da}

\section{Related Work}
\input{paper/8_related_work.tex}

\section{Conclusion}
\input{paper/9_conclusion.tex}

% \clearpage
\section*{Limitations}
\input{paper/limitations}

% \section*{Ethics Statement}
% \input{paper/ethics_and_broader_impact}

% \section*{Acknowledgements}
% This document has been adapted by Jordan Boyd-Graber, Naoaki Okazaki, Anna Rogers from the style files used for earlier ACL, EMNLP and NAACL proceedings.

\bibliography{acl2023}
\bibliographystyle{acl_natbib}

\input{paper/appendix}

\end{document}

%% file: paper/1_abstract.tex
\textit{Semi-supervised learning} (\ssl) is a popular setting aiming to effectively utilize unlabelled data to improve model performance in downstream natural language processing (\nlp) tasks.
Currently, there are two popular approaches to make use of unlabelled data: \textit{Self-training} (\st) and \textit{Task-adaptive pre-training} (\tapt).
\st uses a teacher model to assign pseudo-labels to the unlabelled data, while \tapt continues pre-training on the unlabelled data before fine-tuning. 
To the best of our knowledge, the effectiveness of \tapt in \ssl tasks has not been systematically studied, and no previous work has directly compared \tapt and \st in terms of their ability to utilize the pool of unlabelled data. 
In this paper, we provide an extensive empirical study comparing five state-of-the-art \st approaches and \tapt across various \nlp tasks and data sizes, including in- and out-of-domain settings. 
Surprisingly, we find that \tapt is a strong and more robust \ssl learner, even when using just a few hundred unlabelled samples or in the presence of domain shifts, compared to more sophisticated \st approaches, and tends to bring greater improvements in \ssl than in fully-supervised settings. 
Our further analysis demonstrates the risks of using ST approaches when the size of labelled or unlabelled data is small or when domain shifts exist.
We offer a fresh perspective for future \ssl research, suggesting the use of unsupervised pre-training objectives over dependency on pseudo labels.\footnote{~Code is available at \url{https://github.com/amzn/pretraining-or-self-training}.}

%% file: paper/2_introduction.tex
Pre-training (\pt) language models (LMs) \cite{devlin2018bert,liu2019roberta,radford2019language} over large amounts of text data (e.g. with masked language modelling) and then fine-tuning on task-specific labelled data offer large performance gains across \nlp tasks. 
\textit{Semi-supervised learning} (\ssl)~\cite{grandvalet2004semi,chapelle2009semi,kipf2017semi} is a powerful and effective approach to utilize unlabelled data.
A typical \ssl setting assumes access to a (relatively small) labelled training set and an (often large) unlabelled set. The goal of \ssl is to make effective use of the unlabelled data to improve model (i.e. LMs) performance.

In \nlp, \textit{Self-training} (\st) approaches have been proposed to produce pseudo labels for unlabelled examples to train the model \cite[e.g.\ in][]{yarowsky-1995-unsupervised,mcclosky-etal-2006-effective}. With the advent of neural networks, \st approaches typically focus on using student-teacher models to assign pseudo-labels to the unlabelled data \cite[e.g.\ in][]{artetxe-etal-2018-robust,cai-lapata-2019-semi,dong-de-melo-2019-robust,10.5555/3495724.3496249,gera2022zero}.
Apart from the sophisticated \st approaches, \citet{gururangan-etal-2020-dont} proposed \textit{task adaptive pre-training} (\tapt), which is a straightforward yet effective method for utilising unlabelled examples. This method involves continuing pre-training the LM on the task-specific data without using labels, before proceeding with fully-supervised fine-tuning.
\tapt and \st are both motivated by the need for effectively leveraging unlabelled examples, raising the questions of how TAPT performs in \ssl tasks, as well as how these two approaches perform against each other.

In this work, we investigate the performance of \tapt against five state-of-the-art \st approaches across five \nlp tasks (\cref{sec:st_vs_tapt}). We empirically show that \tapt outperforms all state-of-the-art \st approaches on several tasks, suggesting that it should serve as a strong baseline for \ssl methods. 
Previous research \cite{gururangan-etal-2020-dont} has shown that TAPT can improve performance in fully-supervised settings. Our study goes further by showing that TAPT can be even more effective in \ssl settings (\cref{paragraph:42}).

We next study the impact of using different amounts of labelled and unlabelled data for \ssl (\cref{sec:explore_limits}). Our experiments show that \st approaches are prone to suffering from insufficient labelled or unlabelled data, while \tapt is more robust across different combinations of labelled and unlabelled data sizes. Contrary to the common assumption that \tapt requires a large amount of data to perform well \cite[e.g.][]{li-etal-2021-task-adaptive,10.1162/tacl_a_00517}, our results show that \tapt improves performance with just a hundred unlabelled samples.
We conduct further analysis on the impact of domain shifts in labelled or unlabelled data. While \st approaches generally suffer from domain shifts, \tapt is more robust and even benefits from domain shifts (\cref{sec:da}).

In summary, the main contributions of this paper are as follows:
\begin{itemize}
    \item An extensive empirical study to directly compare five state-of-the-art \st approaches and \tapt across various \nlp tasks in \ssl, with varying amounts of labelled and unlabelled data as well as the effect of domain shifts;
    \item Practical insights learned about the limitations of \st approaches, alongside an exploration of the often-unrecognized yet impressive capacity of  \tapt as a simple, stable and powerful \ssl learner;
    \item A fresh perspective for future \ssl research by demonstrating that leveraging unsupervised signals from unlabelled texts presents a promising and effective approach alternative to dependence on pseudo labels. 
\end{itemize}

%% file: paper/3_background.tex
\subsection{Task Adaptive Pre-training (\tapt)} 
LMs are adapted to downstream \nlp tasks by fine-tuning (\ft) on task-specific data.  \tapt introduces a simple additional step before fine-tuning by continuing pre-training with a masked language modelling (\mlm) objective~\cite{devlin2018bert,liu2019roberta} on the task-specific data without requiring labels.
The main advantage of \tapt is that it provides a simple way for the LM to explore the task space while it can easily make use of all available labelled and unlabelled data. 

\subsection{Self-training (\st)} 
% We provide a basic overview of the \st framework. 
The core idea behind \st approaches is to utilise a teacher model trained on labelled examples to make predictions for unlabelled examples, and train a new student model with these predictions. 
Formally, let $L \triangleq \{(x_1, y_1), \ldots, (x_n, y_n)\}$ denote $n$ labelled examples and $U \triangleq \{\tilde{x_1}, \ldots, \tilde{x_m}\}$ denote $m$ unlabelled examples, where usually $m \gg n$. % :We consider classification and assume the data is partitioned into $K$ classes. 
The \st framework is trained with three main steps as follows.

\paragraph{Step 1.} A teacher model $F$, parameterized by a neural network $\Theta$, is trained via minimizing the cross entropy loss $\ell$ on labelled examples $L$:
\begin{equation}
    \label{equation:step_1}
    \mathcal{L}_{teacher}(L) = \sum \limits_{x_i, y_i \in L} \ell(y_i, F(x_i, \Theta)),
\end{equation}
\input{table/datasets}

\paragraph{Step 2.} The teacher model $F$ is used to make predictions (referred to as “pseudo-labels”) on unlabelled examples $U$:
\begin{equation}
    \label{equation:step_2}
    \tilde{y_i} = F(\tilde{x_i}, \Theta),
\end{equation}
where $\tilde{y_i}$ can be either the continuous logit or the discrete label induced by an \textsc{ArgMax} operation.

\paragraph{Step 3.} A student model $G$, parameterized by a fresh neural network $\Phi$, is trained to fit labelled and pseudo-labelled examples:
\begin{multline}
    \label{equation:step_3}
    \mathcal{L}_{student}(L, U) = 
        \sum \limits_{x_i, y_i \in L} \ell(y_i, F(x_i, \Phi)) \\
        + \sum \limits_{\tilde{x_i}, \tilde{y_i} \in U} \ell(\tilde{y_i}, F(\tilde{x_i}, \Phi))
\end{multline}

This process is repeated for a given number of times by treating the student as a new teacher to re-predict pseudo-labels as in \cref{equation:step_2} and then training a new student with \cref{equation:step_3}. In practice, \st with techniques such as consistency regularization \cite{miyato2018virtual,clark-etal-2018-semi,berthelot2019mixmatch}, strong data augmentation \cite{sohn2020fixmatch,xie2020self,10.5555/3495724.3496249}, confidence threshold \cite{sohn2020fixmatch,zhang2021flexmatch,berthelot2021adamatch} usually leads to substantial improvements in model performance.

%% file: table/datasets.tex
\begin{table*}[hbt!]
\centering
\begin{adjustbox}{max width=\textwidth}
\begin{tabular}{llrrrrcc}
\toprule
\bf Dataset                                 & \bf Task Type                  & \bf Train Size & \bf Dev. Size  & \bf Test Size  & \bf $|\mathcal{Y}|$ & \bf $L$  \\ \midrule
\imdb \cite{maas-etal-2011-learning}        & Movie Review Sentiment         & 23,000         & 2,000          & 25,000         & 2         &  149 \\ % 8H
\sst \cite{wang-etal-2018-glue}            & Movie Review Sentiment          & 60,000         & 7,349          & 872            & 2         & 37 \\
\ag \cite{10.5555/2969239.2969312}         & News Topic Classification       & 100,000        & 10,000         & 7,600          & 4         & 134\\ % 20H
\amazon \cite{10.1145/2507157.2507163}     & Product Review Sentiment        & 250,000        & 25,000         & 650,000        & 5         & 79 \\
% \yelp         & Product Review Sentiment       & 250,000        & 25,000         & 50,000         & 5          \\
\yahoo \cite{chang2008importance}         & Topic Classification             & 500,000        & 50,000         & 60,000         & 10        & 32 \\ 
\bottomrule
\end{tabular}
\end{adjustbox}
\caption{Statistics of datasets. $|\mathcal{Y}|$: \# of classes for classification tasks. $L$: average \# of words in input sentence(s). Note that we only sample examples from the original training set in our experiments.}
\label{table:datasets}
\end{table*}

%% file: paper/4_setup.tex
\paragraph{Datasets.} We experiment with five datasets used in previous related work for \ssl \cite{gururangan-etal-2019-variational,chen-etal-2020-mixtext,10.5555/3495724.3496249,li-etal-2021-semi-supervised,gera2022zero}, including \imdb \cite{maas-etal-2011-learning}, \sst \cite{wang-etal-2018-glue}, \ag \cite{10.5555/2969239.2969312}, \amazon \cite{10.1145/2507157.2507163}, 
and \yahoo \cite{chang2008importance}. Table \ref{table:datasets} shows data statistics. We also provide descriptions and examples of datasets in Appendix \sect{appendix:description}. We show the process for quantifying the similarity between datasets in Appendix \sect{appendix:dataset_similarity}. Adhering to previous work \cite[e.g.][]{chen-etal-2020-mixtext,wang2022usb}, we sample the same amount of labelled data per class from the train set, given the labelled size, to form the labelled set. We re-sample the labelled data using the same five seeds for all different approaches and report the average performance with an error bar.

\paragraph{\tapt.} % Following the setting in the previous work \cite{gururangan-etal-2020-dont}, 
Our approach to \textit{task adaptive pre-training} (\tapt) using \robertabase \cite{liu2019roberta} is to further pre-train on the training text corpus including labelled and unlabelled data (see Table \ref{table:tapt_hyperparameters} in Appendix for hyperparameter details). % The \tapt size is equivalent to the unlabelled data size. We use the term `\tapt size' to refer to the sample size of the unlabelled training text corpus. 
The model is then fine-tuned on the labelled data where the \texttt{[CLS]} token representation is passed to an extra feed-forward layer for classification (see Table \ref{table:finetune_hyperparameters} in Appendix for hyperparameter details). 
The process of \tapt + \textsc{Fine-tuning} is simply denoted by \tapt henceforth.

\paragraph{\st.} We implement five state-of-the-art \st approaches, including VAT \cite{miyato2018virtual}, FixMatch \cite{sohn2020fixmatch}, Dash \cite{xu2021dash}, FlexMatch \cite{zhang2021flexmatch}, and AdaMatch \cite{berthelot2021adamatch} (see descriptions of these approaches in Appendix \sect{appendix:baseline_models}). 
We use \robertabase as the backbone, and the \texttt{[CLS]} token representation with an extra feed-forward layer is used for classification (see Table  \ref{table:semi_hyperparameters} in Appendix for hyperparameter details). Adhering to previous work \cite{10.5555/3495724.3496249,wang2022usb}, back-translation \cite{ott2019fairseq} is used for data augmentation.

\paragraph{Baselines.} For reference, we also evaluate two baseline models that are only fine-tuned (from an off-the-shelf \robertabase checkpoint) on: (1) the same labelled set as \tapt and \st (\partials); and (2) the whole training set (\fullys).

%% file: paper/5_st_vs_tapt.tex
\input{table/experiments}
\begin{figure*}[!ht]
  \centering
  \includegraphics[width=\textwidth]{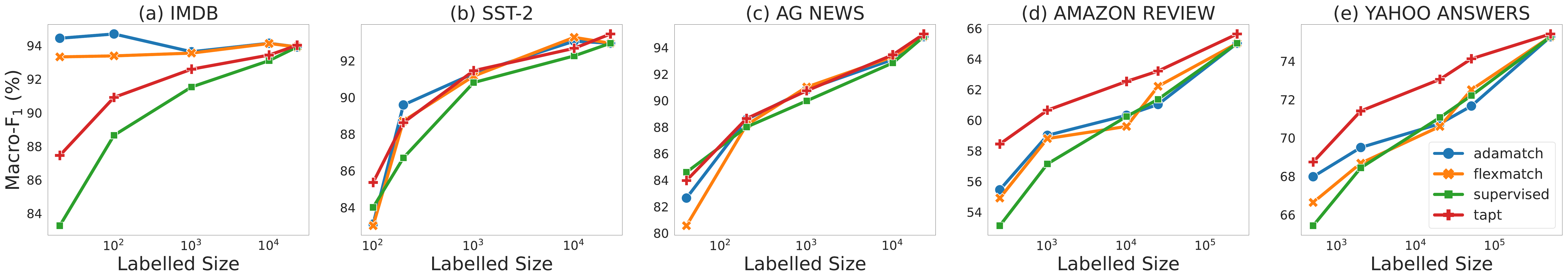}
  \caption{The effect of labelled size on \tapt and \st. Average test Macro-$F_1$ score over 5 seeds is reported. From the left to the right, \tapt and \st utilizes $23$k, $60$k, $100$k, $250$k, and $500$k unlabelled samples respectively.}
  \label{fig:f1_wrt_labelled_sample_size}
\end{figure*}

\paragraph{Overview.} \label{paragraph:overview} 
Table \ref{table:main_results} shows the performance of \tapt against five state-of-the-art \st approaches and the baselines (\partials and \fullys) across five datasets, each with two different sizes of labelled data for training following \citet{wang2022usb}. 
Overall, we observe that:
(\hyperref[paragraph:41]{\color{red}1}) \tapt achieves highly competitive results compared with state-of-the-art \st approaches; and
(\hyperref[paragraph:42]{\color{red}2}) \tapt gains more improvement compared to the \partials baselines when using fewer labelled samples. 

For our first finding, the experimental results show that \tapt outperforms all five state-of-the-art \st approaches with lower variances on \amazon, and \yahoo, as shown in Table \ref{table:main_results}. For example, \tapt obtains a $F_1$ score of 68.8\% compared to the best \st approach's $F_1$ score of 68.0\% (using 500 labelled samples) and 71.5\% compared to \st's 69.6\% (using 2000 labelled samples) on \yahoo. 
For an example of the second finding, \tapt gains 3.6\% $F_1$ improvement over \partials (using 20 labelled samples) compared to 2.2\% (using 100 labelled samples) on \imdb. Below we delve deeper into these two findings and discuss them in more detail.

\paragraph{\#1. \tapt is a strong semi-supervised learner and can outperform state-of-the-art \st approaches.} 
\label{paragraph:41}
Figure \ref{fig:f1_wrt_labelled_sample_size} shows how the performance of \st, \tapt, and \partials vary with respect to five different labelled sizes on each dataset, where two latest \st approaches (\adamatch and \flexmatch) are selected as representatives for \st.
Experimental results further verify that \tapt has a consistent advantage over \adamatch and \flexmatch across different labelled sizes on \amazon and \yahoo. It is also worth noting that, while \tapt brings a stable improvement over \partials across all datasets with varying labelled sizes, \st can sometimes bring more substantial improvement, for example when only a few hundreds of labelled samples are available from \imdb. However, we do not observe similar phenomena for \st on other datasets.
Our experimental results demonstrate that \tapt is a simple, effective and strong learner for \ssl tasks, and it should serve as a baseline for \ssl tasks in \nlp.

\begin{figure}[!t]
  \centering
  \includegraphics[width=0.5\textwidth]{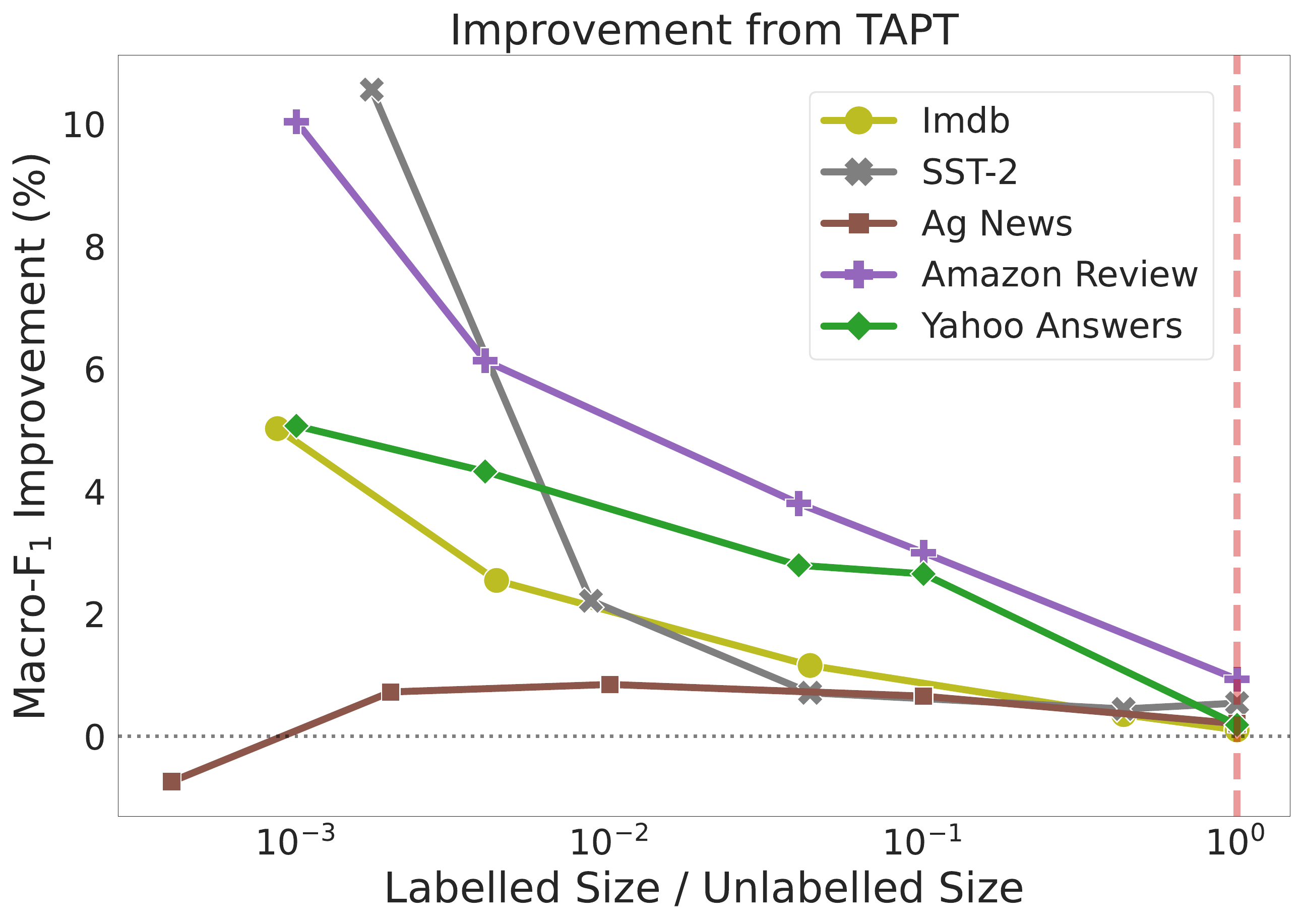}
  \caption{The impact of labelled size on the $F_1$ improvement from \tapt over \partials, where unlabelled size is fixed for each dataset. The red vertical line highlights the \fullys setting on which prior work \cite{gururangan-etal-2020-dont} focuses.}% {\color{red} Add similar figure for \st in Appendix.}{
  \label{fig:f1_wrt_tapt_5}
\end{figure}
\begin{figure*}[!ht]
  \centering
  \includegraphics[width=\textwidth]{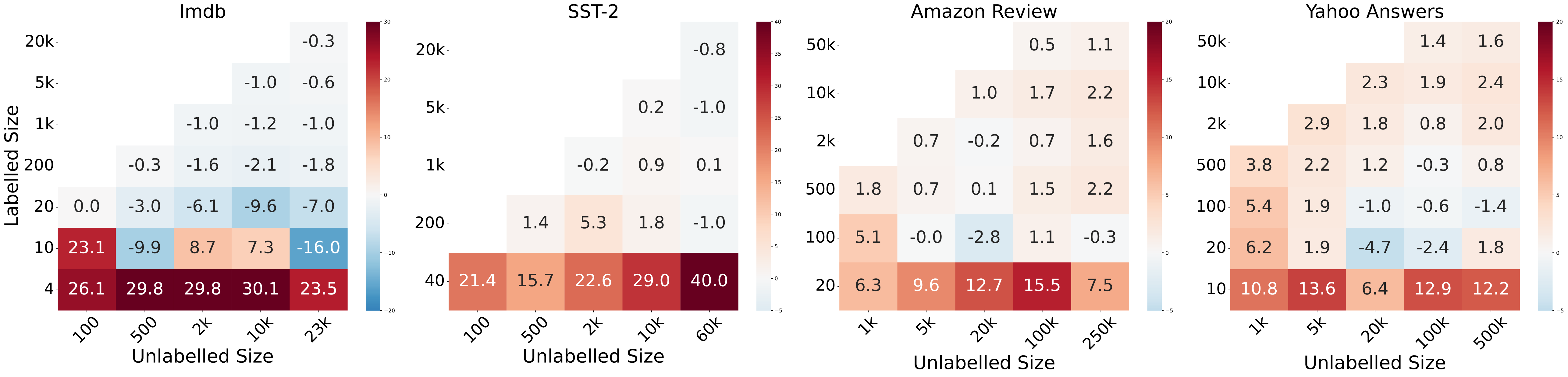}
  \caption{Performance difference between \tapt and \st with varying labelled and unlabelled sizes on \imdb, \sst, \amazon and \yahoo. Positive values indicate that \tapt performs better, while negative values indicate that \st performs better. Average Macro-$F_1$ score on test sets over five seeds is reported.}
  \label{fig:heatmap}
\end{figure*}

\paragraph{\#2. \tapt tends to bring more improvements in \ssl than in \fullys setting.} 
\label{paragraph:42}
We further study the behaviour of \tapt \textit{itself} under \ssl, where we select \partials as the baseline rather than \st approaches.
Figure \ref{fig:f1_wrt_labelled_sample_size} shows that the differences in performance (in absolute values) between \tapt (red lines) and \partials (green lines) generally increase as the labelled size decreases.
To gain a better understanding of the impact of labelled data sizes, we plot the improvement from \tapt over \partials (in percentages) against the ratio between labelled size and unlabelled size (unlabelled size is fixed for each dataset) in Figure \ref{fig:f1_wrt_tapt_5}. We see that \tapt improves over \partials further as the ratio of labelled and unlabelled sizes decreases, highlighting the trends of gaining greater improvement in low-resource \ssl setting. This finding is complementary to prior works \cite[e.g.\ in][]{howard-ruder-2018-universal,gururangan-etal-2020-dont} that focus on \tapt's improvement from the \fullys perspective, represented by the rightmost red vertical line in Figure \ref{fig:f1_wrt_tapt_5}.
The rising trend of the improvement is not monotonic as the labelled size is reduced.
Rather it could provide insight into how \tapt improves over \partials in \ssl and inspire the design of new approaches.

%% file: table/experiments.tex
\begin{table*}[t!]
\centering
% \small
% \begin{adjustbox}{max width=\textwidth}
\resizebox{\textwidth}{!}{
\begin{tabular}{lccccccccccc}
% \begin{tabular}{lCCCCCCCCCC}
\toprule
\multirow{2}{*}{\bf Method} & \multicolumn{2}{c}{\bf \imdb}      & \multicolumn{2}{c}{\bf \sst}     & \multicolumn{2}{c}{\bf \ag}     & \multicolumn{2}{c}{\bf \amazon}   & \multicolumn{2}{c}{\bf \yahoo}
\cr                         \cmidrule(lr){2-3}                   \cmidrule(lr){4-5}                \cmidrule(lr){6-7}                 \cmidrule(lr){8-9}               \cmidrule(lr){10-11}                           
                            & 20              & 100              & 40             & 100              & 40           & 200              & 250           & 1000                    & 500          & 2000     \\ 
\midrule
\multicolumn{11}{l}{\bf \st Approaches}  \\
\vat                        & 90.2$_{0.9}$    & 92.0$_{0.4}$.    &\se75.0$_{12.0}$&\hl86.2$_{3.4}$    &\hl87.5$_{1.0}$ &\hl89.5$_{0.7}$ & 52.2$_{1.3}$  & 57.5$_{0.2}$       & 66.9$_{0.5}$ & 68.6$_{0.2}$  \\
\fixmatch                   &\se93.4$_{0.1}$  & 93.4$_{0.1}$     & 37.3$_{8.5}$   &  66.4$_{21.3}$    & 75.6$_{8.7}$ &\se88.8$_{0.6}$   &\se55.9$_{1.1}$& 59.0$_{0.5}$        & 67.5$_{1.0}$ &\se69.6$_{0.4}$  \\
\dash                       & 93.2$_{0.3}$    & 93.4$_{0.2}$     & 38.2$_{10.1}$  & 73.3$_{18.6}$   & 74.3$_{6.6}$ &  88.5$_{0.6}$    & 56.6$_{1.8}$  & 59.3$_{0.2}$      & 67.6$_{1.0}$ & 69.5$_{0.3}$   \\
\flexmatch                  & 93.3$_{0.1}$    &\se93.4$_{0.1}$   & 40.6$_{7.7}$   &  83.0$_{8.3}$    & 80.6$_{4.4}$ &  88.2$_{0.5}$    & 54.9$_{3.9}$  & 58.8$_{0.4}$        & 66.6$_{0.7}$ & 68.7$_{0.4}$   \\
\adamatch                   &\hl94.4$_{0.4}$. &\hl94.7$_{0.2}$   & 42.6$_{13.3}$  & 83.1$_{4.4}$   & 82.7$_{5.9}$ &  88.6$_{0.4}$    & 55.5$_{2.8}$  &\se59.0$_{0.7}$     &\se68.0$_{0.7}$& 69.5$_{0.3}$   \\
\midrule
\partials                   & 83.3$_{7.4}$  & 88.7$_{0.2}$       & 74.7$_{6.1}$   & 84.0$_{2.7}$   &\se84.6$_{1.6}$& 88.0$_{0.8}$    & 53.1$_{0.7}$  & 57.2$_{0.1}$       & 65.4$_{0.3}$ & 68.5$_{0.3}$ \\
\quad + \tapt               & 86.9$_{2.8}$  & 90.9$_{0.6}$       &\hl 82.6$_{4.0}$&\se 85.4$_{2.4}$  & 84.0$_{1.3}$&  88.7$_{0.7}$     &\hl58.4$_{0.7}$&\hl60.6$_{0.1}$  &\hl68.8$_{0.7}$&\hl71.5$_{0.3}$\\
\midrule
\fullys                    &\multicolumn{2}{c}{93.9$_{0.1}$}    & \multicolumn{2}{c}{93.0$_{0.6}$} & \multicolumn{2}{c}{94.8$_{0.1}$} & \multicolumn{2}{c}{65.0$_{0.2}$} & \multicolumn{2}{c}{75.3$_{0.2}$} \\
\quad + \tapt               &\multicolumn{2}{c}{94.0$_{0.2}$}    & \multicolumn{2}{c}{93.5$_{0.3}$} &\multicolumn{2}{c}{95.0$_{0.1}$}   &\multicolumn{2}{c}{65.6$_{0.1}$}   &\multicolumn{2}{c}{75.4$_{0.1}$} \\
\bottomrule
\end{tabular}
}
% \end{adjustbox}
\caption{Performance of \tapt, \st approaches and the baselines across five datasets using two different sizes of the training labelled data. We report average Macro-$F_1$ on the test set across five seeds, with standard deviations in subscripts. Blue and orange represent the best and second-best performance in a column respectively.}
\label{table:main_results}
\end{table*}

%% file: paper/6_analysis.tex
In \cref{sec:st_vs_tapt}, our experimental results showed inconsistent results across datasets. For example, \st performs better on \imdb while \tapt achieves better results on \amazon and \yahoo. 
We hypothesize that this might be attributed to the exposure to different sizes of labelled or unlabelled data. 
To verify this hypothesis and shed light on the differences in performance between datasets,
we compare \tapt and \st (using \adamatch and \flexmatch as representatives) by sampling different labelled and unlabelled sizes in \imdb, \sst, \amazon and \yahoo.

Figure \ref{fig:heatmap} visualizes the differences in performance between \tapt and \st, where each cell represents the macro-$F_1$ performance difference of \tapt over \st (averaged across five seeds). In each case, the highest performance among \flexmatch and \adamatch is selected to represent the performance of \st.
Overall, we observe that:
(\hyperref[paragraph:51]{\color{red}1}) \tapt improves the fine-tuning performance even with a few hundred unlabelled examples; and
(\hyperref[paragraph:52]{\color{red}2}) \tapt performs more stable across the different labelled and unlabelled data sizes than \st approaches. 
Below we provide a comprehensive analysis of the impact of labelled and unlabelled sizes.

\begin{figure}[!t]
  \centering
  \includegraphics[width=\columnwidth]{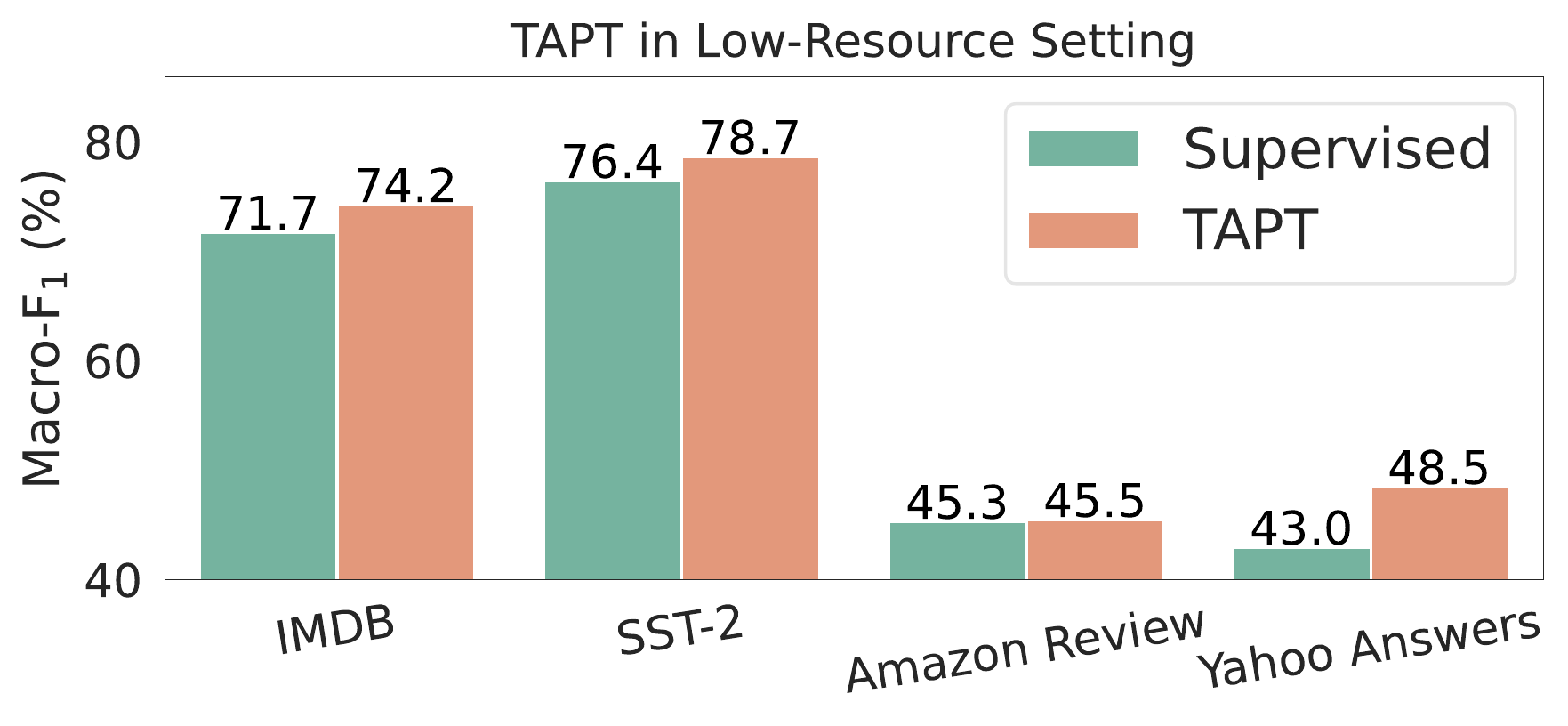}
  \caption{The performance of \tapt against the \partials baseline in the low-resource setting of unlabelled data. From the left to the right, \tapt utilizes $100$, $100$, $1\,000$, and $1\,000$ unlabelled samples respectively.}
  \label{fig:tapt_low_resource}
\end{figure}

\begin{figure}[t!]
  \centering
  \includegraphics[width=\columnwidth]{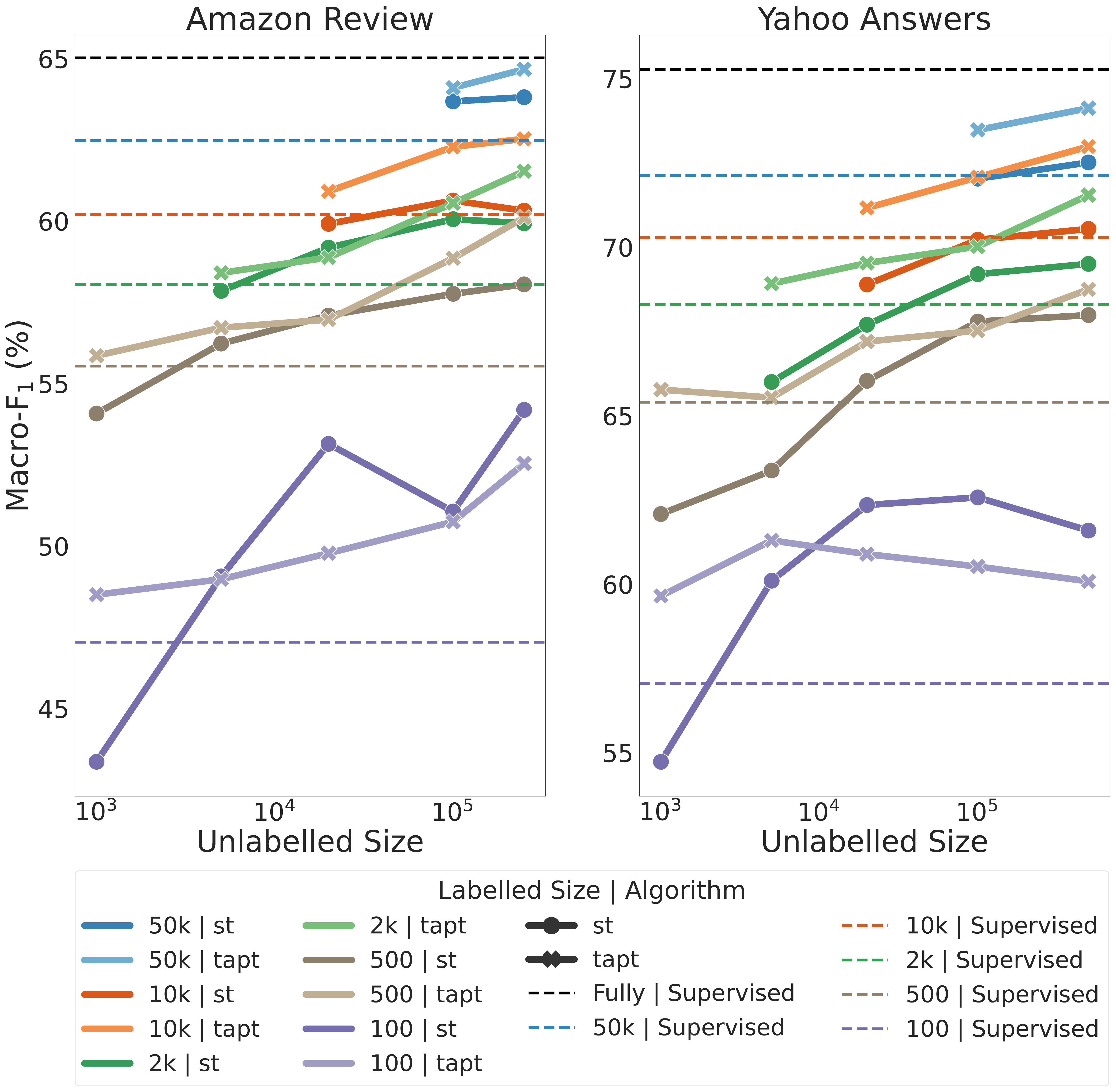}
  \caption{The performance of \st and \tapt using different unlabelled sizes. Average test results across five seeds are reported where the best result from \flexmatch and \adamatch is selected to represent \st.}
  \label{fig:unlabelled_size_impact}
\end{figure}

\paragraph{\#1. \tapt works even with a few hundred unlabelled samples.} \label{paragraph:51}
It is generally assumed that \tapt requires a large amount of unlabelled data to perform well \cite[e.g.][]{li-etal-2021-task-adaptive,10.1162/tacl_a_00517}. However, we surprisingly observe that \tapt can bring substantial improvement over \partials baseline even with a relatively small number of unlabelled samples, as shown in Figure \ref{fig:unlabelled_size_impact}. 
To explore the effectiveness of \tapt over \partials in the low-resource setting of unlabelled data, we select the performance of \tapt and \partials from the first column (the lowest unlabelled size) for each dataset in Figure \ref{fig:heatmap} and plot their average performance over different labelled sizes. Figure \ref{fig:tapt_low_resource} shows that \tapt improves over the \partials baseline with just one hundred or one thousand samples. For instance, \tapt achieves a 5.5\% increase in $F_1$ score compared to the \partials baseline when using only 1k unlabelled samples on \yahoo.
Additionally, this performance is achieved without the need for large amounts of tokens in each sample, as training samples from \sst, on average, contain only 9 tokens and training samples from \yahoo contain about 32 tokens (see examples in Table \ref{table:dataset_example} of Appendix).

\paragraph{\#2. Scarce labelled data and adequate unlabelled data.}
\label{paragraph:52}
\tapt appears to be a more favourable choice than \st approaches in this setting.
The bottom of each sub-figure in Figure \ref{fig:heatmap} shows a clear labelled size boundary, below which \flexmatch and \adamatch are outperformed by \tapt with a large margin, regardless of datasets and unlabelled size used. This suggests that \st might not be able to efficiently handle large amounts of unlabelled data if labelled data do not provide adequate information. This might be attributed to \textit{confirmation bias} \cite{tarvainen2017mean,arazo2020pseudo}, which results from the accumulation of errors in the iterative \st process caused by incorrect pseudo-labels.

The specific value of adequate labelled size boundary for \st approaches depends on the nature of the dataset. 
For example, even though both \imdb and \sst are binary classification tasks for movie review sentiment analysis, the labelled size boundary for \sst is higher ($40>4$), indicating that this boundary tends to increase as the task becomes more challenging. While it may be easy to obtain dozens of labelled data in this case, when the task becomes more intricate or contains noisy weak labels, it is important to be aware of this potential issue with \st approaches. \tapt could serve as an alternative in situations where collecting adequate labelled data for training is costly. We provide specific values of the performance of \st and \tapt, and further verify that this finding applies to other \st approaches in Appendix \sect{sec:verification}.

\paragraph{\#3. Adequate labelled data and scarce unlabelled data.} 
\label{paragraph:unlabel_initialize}
In this setting, \tapt is more robust, while \st has a greater chance of performing worse than the \partials baseline.
In Figure \ref{fig:unlabelled_size_impact}, we plot the performance of \st approaches and \tapt against five different sizes of unlabelled data, grouped by size (using similar colours). We note that \st approaches perform worse than their corresponding \partials baselines (represented by horizontal lines) until a certain amount of unlabelled data has been reached. For example, when the labelled size is 500, \st requires about 20k unlabelled samples to achieve the corresponding \partials baseline performance on \yahoo. On the other hand, \tapt generally outperforms \partials baselines demonstrating its robustness across various unlabelled sizes.

To further quantify the model performance in case of scarce unlabelled and adequate labelled data, we choose the three lowest unlabelled sizes (the first three columns) excluding the lowest labelled size (the last row) in Figure \ref{fig:heatmap} for each dataset. Our analysis shows that \st has 67\%, 56\% and 54\% probability of falling below the \partials baselines on \sst, \amazon, and \yahoo respectively. Even on \imdb where \st generally performs well, it still has a probability of 33\% to fall behind \partials. In contrast, \tapt never performs worse than \partials in those cases. We provide computation details and comparative statistics in Appendix \sect{sec:falling_below_baseline}.

The specific value of adequate unlabelled size boundary for \st approaches depends on the nature of the dataset as well as the labelled size.
Figure \ref{fig:unlabelled_size_impact} illustrates that as the size of the labelled data increases, \st approaches require more unlabelled data to surpass the \partials baselines. For example, on \amazon, \st trained with 100 labelled samples requires about 5k unlabelled samples to perform better than \partials, while \st trained with 10k labelled samples requires about 100k unlabelled samples. Adjusting the unlabelled size accordingly might be conducive to exploiting the full potential of \st approaches.

\input{table/yahoo_low_labeled_low_unlabel}
\begin{figure*}[!ht]
  \centering
  \includegraphics[width=\textwidth]{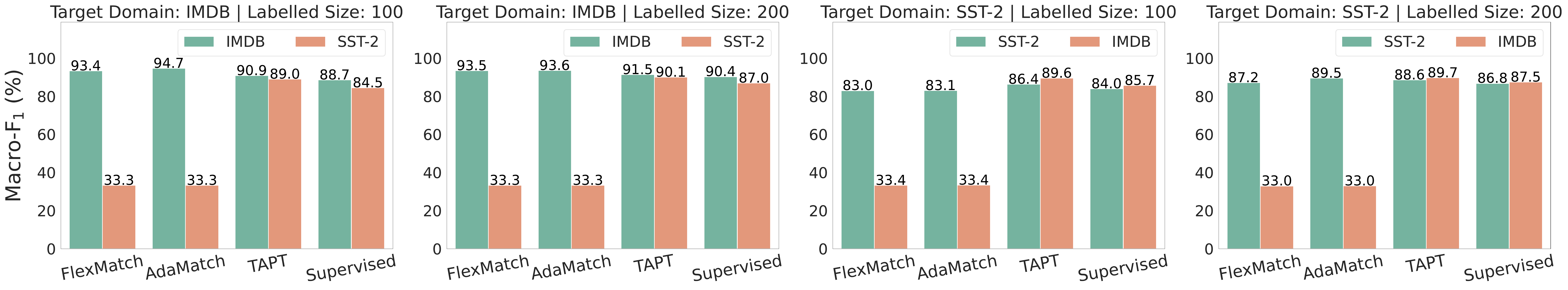}
  \caption{Results of \uda experiments. Legends indicate domains of labelled training data. Orange/green represents the performance with/without domain shift. Average Macro-$F_1$ score on test sets over five seeds is reported.}
  \label{fig:uda}
\end{figure*}
\input{table/domain_adaptation_noise_unlabel}

\input{table/domain_adaptation_categories}

\paragraph{\#4. Scarce labelled and unlabelled data.}
\label{paragraph:case_3}
When the labelled data is insufficient, increasing unlabelled size is not helpful or even detrimental to \st approaches. 
This finding is 
% consistent with the findings from prior work \cite{wang2022usb} and 
well-illustrated in the last row of results on \sst shown in Figure \ref{fig:heatmap}. In other words, reducing the size of unlabelled data could be beneficial for \st approaches when the labelled size is inadequate. We further zoom in on this phenomenon in Table \ref{table:yahoo_low_labeled_low_unlabel} by selecting 4 fixed labelled and 500 unlabelled samples, and gradually removing unlabelled samples on \imdb.
This is a stark contrast to the case where more unlabelled data is beneficial for \st approaches when adequate labelled data is available. Meanwhile, \tapt generally benefits from training on more in-domain unlabelled data, following the scaling law in LMs \cite{kaplan2020scaling,hoffmann2022training}.

\paragraph{\#5. Adequate labelled and unlabelled data.}
Both \st and \tapt have demonstrated the ability to exploit unlabelled data in this setting.
Figure \ref{fig:heatmap} shows that \st dominates in \imdb when more than 10 labelled and 100 unlabelled samples are available. On the other hand, \tapt generally performs better than \st on \amazon and \yahoo, indicating that
the answer to which approach is better depends on the nature of the dataset and task.
As labelled and unlabelled data size increase, the difference between \st and \tapt shrinks (colours fade and lines converge in Figures \ref{fig:heatmap} and \ref{fig:unlabelled_size_impact}). 
As the labelled data in size reaches the unlabelled data, the method of \st reduces to \fullys, which is generally outperformed by \tapt \cite{gururangan-etal-2020-dont}. 

%% file: table/yahoo_low_labeled_low_unlabel.tex
\begin{table}[t!]
\centering
% \footnotesize
% \scriptsize
\begin{adjustbox}{max width=\columnwidth}
\begin{tabular}{lcccc}
\toprule
                %   & \multicolumn{4}{c}{\bf Unlabelled Size} \\ 
        %   \cmidrule{2-5} 
\bf \#Unl.             & \bf 10                & \bf 50               & \bf 100                 & \bf 500  \\ 
\midrule
\flexmatch             &  57.3$_{17.9}$        & 35.2$_{3.4}$         &  45.1$_{22.5}$          &  33.4$_{0.1}$  \\
\adamatch              &  53.3$_{22.1}$        & 36.8$_{6.1}$         &  33.5$_{0.2}$           &  33.6$_{0.3}$   \\
\bottomrule
\end{tabular}
\end{adjustbox}
\caption{Test results on \imdb with 4 fixed labelled data. An average Macro-$F_1$ score over five seeds is reported.}
\label{table:yahoo_low_labeled_low_unlabel}
\end{table}

% IMDB
%                 1.        2.        3.       4.         5.          Mean.     Std.
% adamatch_10     34.76385, 37.60069, 79.38618,33.4168,   81.140523,  53.3$_{22.1}$
% flexmatch_10    56.53706, 62.76019, 45.94878,34.006386, 87.316219,  57.3137$_{17.9043}$
%
% adamatch_50     34.481727,33.3511,  49.03275,33.42216, 33.705583,    36.8$_{6.1}$
% flexmatch_50    33.37598, 33.3955253,33.80272,33.33333,41.955613,    35.1726$_{3.3957}$
%
% adamatch_100   33.758585, 33.36887, 33.59059, 33.342221,33.54631,   33.5$_{0.2}$
% flexmatch_100. 33.3333,   33.34222, 35.30271, 33.3422,  90.11407,   45.0869$_{22.5264}$

% Yahoo
%                 1.        2.        3.       4.         5.          Mean.     Std.
% adamatch_20     15.23896, 14.02273, 3.066095,24.5670,   28.22,
% flexmatch_20    11.0952,  12.32139, 9.3850,  22.23198,  32.272957,
% 14.6722, 22.9138, uswest3
%
% adamatch_100    4.457576,3.97081,  3.772360,8.46781,    20.923124,
% flexmatch_100   12.12229,8.01257,  2.56040, 8.33811,    23.464524, uswest7
%
% adamatch_1000    21.04469, 10.97971,9.12794,26.47056,   6.13947,
% flexmatch_1000   12.0486,13.83794, 15.0674, 12.228179,  18.060973,
%
% adamatch_10000    25.4960,4.94542,16.33674, 13.834,     12.46424,
% flexmatch_10000   3.384770,3.11006,5.42428, 2.50538,     3.861112,
%
%
% adamatch_100000   20.78733,7.14495,4.8340, 16.19207,    10.1916,
% flexmatch_100000  1.8288,  2.04729,1.84127, 1.835,      1.845060,

%% file: table/domain_adaptation_noise_unlabel.tex
% =========================================
% Cross-domain results
% =========================================
\newcommand{\bsl}{\makebox[0pt][r]{\raisebox{-0.05em}{$\bigstar\,$}}}

\begin{table*}[hbt!]
\centering
\scriptsize
% \footnotesize
% \setlength\tabcolsep{2pt}
\resizebox{\textwidth}{!}{
\begin{tabular}{lllllll}
\toprule
\bf Train (Lab.)          & \bf Train (Unl.) & \bf \#Lab. & \bf \flexmatch          & \bf \adamatch            & \bf \tapt              & \bf \partials  \\
\midrule
% \multirow{3}{*}{\imdb}    & \imdb           & 20         & 93.3$_{0.1}$            & \hl94.4$_{0.4}$          & 86.9$_{2.8}$           & 83.3$_{7.4}$ ~~~~~$\bigstar\,$\\
%                           & \sst            & 20         & 79.0$_{3.4}$\down{15.3} & 71.1$_{12.6}$\down{24.7} &\hl83.4$_{8.4}$\down{4.0}& 83.3$_{7.4}$ \\
%                           & \amazon         & 20         & 90.9$_{1.0}$\down{2.6}  & \hl92.2$_{0.4}$\down{2.3}& 84.4$_{7.4}$\down{2.9}& 83.3$_{7.4}$ \\
% \midrule
\multirow{3}{*}{\imdb}    & \imdb           & 100        & 93.4$_{0.1}$            & \hl 94.7$_{0.2}$         & 90.9$_{0.6}$            & 88.7$_{0.2}$ ~~~~~$\bigstar\,$\\
                          & \sst            & 100        & 89.1$_{1.2}$\down{4.6}  & 87.6$_{2.2}$\down{7.5}   &\hl89.9$_{0.6}$\down{1.1}& 88.7$_{0.2}$ \\
                          & \amazon         & 100        & 92.1$_{0.7}$\down{1.4}  &\hl92.4$_{0.2}$\down{2.4} & 91.4$_{0.3}$\up{0.6}    & 88.7$_{0.2}$ \\
\midrule
\multirow{3}{*}{\imdb}    & \imdb           & 200         & 93.5$_{0.1}$            & \hl93.6$_{0.1}$          & 91.8$_{0.3}$            & 90.3$_{0.4}$ ~~~~~$\bigstar\,$\\
                          & \sst            & 200         & 89.5$_{2.4}$\down{4.3}  & 88.9$_{1.0}$\down{5.0}   &\hl90.3$_{0.4}$\down{1.6}& 90.3$_{0.4}$ \\
                          & \amazon         & 200         & 92.5$_{0.4}$\down{1.1}  &\hl 92.7$_{0.5}$\down{1.0}& 92.1$_{0.2}$\up{0.3} & 90.3$_{0.4}$ \\
\midrule
\multirow{3}{*}{\sst}     & \sst            & 100        & 83.0$_{8.3}$            & 83.1$_{4.4}$             & \hl 85.4$_{2.4}$       & 84.0$_{2.7}$ ~~~~~$\bigstar\,$\\
                          & \imdb           & 100        & 46.7$_{2.1}$\down{43.7} & 49.2$_{7.3}$\down{40.8}  & \hl88.5$_{0.9}$\up{3.6}& 84.0$_{2.7}$ \\
                          & \amazon         & 100        & 46.4$_{4.9}$\down{44.1} & 48.2$_{11.0}$\down{42.0} & \hl88.9$_{0.9}$\up{4.1}& 84.0$_{2.7}$ \\
\midrule
\multirow{3}{*}{\sst}     & \sst            & 200        & 87.2$_{3.9}$            &\hl89.5$_{0.9}$           & 88.6$_{0.9}$           & 86.8$_{0.3}$ ~~~~~$\bigstar\,$\\
                          & \imdb           & 200        & 62.7$_{7.4}$\down{28.1} & 61.0$_{2.8}$\down{31.8}  &\hl89.1$_{1.1}$\up{0.6} & 86.8$_{0.3}$ \\
                          & \amazon         & 200        & 61.8$_{7.7}$\down{29.1} & 56.0$_{10.3}$\down{17.4} &\hl89.4$_{1.0}$\up{0.9} & 86.8$_{0.3}$ \\
\bottomrule
\end{tabular}
}
\caption{Results of \stl experiments. We report the average Macro-$F_1$ score on the test set across five seeds, with standard deviations as subscripts. Blue represents the best result for each row. Stars highlight rows without domain shifts. Arrows in colours stand for the changes in performances against the star row result within each cell.}
\label{table:suda}
\end{table*}

%% file: table/domain_adaptation_categories.tex
\begin{table}[t!]
\centering
% \footnotesize
% \small
\begin{adjustbox}{max width=\columnwidth}
\begin{tabular}{lll}
\toprule
\bf Task          & \bf Lab.            & \bf Unl.   \\    % & \bf Dev. & \bf Test \\ 
\midrule
\textit{Semi-supervised Learning}             &  Target             & Target     \\    % & target & target  \\
\textit{Unsupervised Domain Adaptation}             &  Source             & Target     \\    % & target & target  \\
% SSDA              &  source + target    & target           \\
% Our Original Plan                          &  target             & source + target  \\
\textit{Self-taught Learning}              &  Target             & Source     \\  % & target & target \\
\bottomrule
\end{tabular}
\end{adjustbox}
\caption{A summary of domain adaptation, where the distribution of source and target domains are different.}
\label{table:da_categories}
\end{table}

%% file: paper/7_da.tex
We next investigate how \st and \tapt compare in the presence of domain shifts between labelled and unlabelled data in two additional settings (refer to Table \ref{table:da_categories}).
First, we experiment with the \textit{Unsupervised Domain Adaptation} (\uda) setting, where domain shifts exist between the labelled data from a source domain and the unlabelled data from the target domain \cite{ben2010theory,saito2018maximum,ramponi_neural_2020}. 
Then, we experiment with \textit{Self-taught Learning} (\stl) \cite{raina2007self} in a domain adaptation setting, where the unlabelled data come from the source domain and the labelled data from the target domain. 
In both settings, we use the (labelled) validation and test sets from the target domain. Validation and test sets are excluded from any pool of labelled or unlabelled train data.

\paragraph{\#1. Unsupervised Domain Adaptation (\uda).} \label{paragraph:uda}
In this setting, we use two movie sentiment datasets, \imdb and \sst, as the source and target domain (and vice versa) with two different sizes of labelled data (i.e. 100 and 200). 

Figure \ref{fig:uda} depicts the performance of \st and \tapt in \uda.
In case of domain shifts, we observe that \flexmatch and \adamatch fail to deliver satisfactory results and their performance drops to the level of random guessing, with a $F_1$ score of 33\% across all labelled sizes and datasets.
This highlights the vulnerability of \st approaches in \uda.
In contrast, \tapt demonstrates robust performance even with domain shifts, on par with its own \ssl performance without domain shifts. 
Additionally, \tapt even benefits from training on the source domain. For instance, training on \imdb (source domain) further improves the performance of \tapt on \sst (target domain) from 86.4\% to 89.6\% with 100 labelled samples and from 88.6\% to 89.7\% with 200 labelled samples.

\paragraph{\#2. Self-taught Learning (\stl).} \label{paragraph:stl}
We select \imdb, \sst, and \amazon for this setting. Although they are all sentiment reviews datasets, \imdb and \amazon are more closely related (see the similarity analysis in Table \ref{table:overlap_imdb_and_amazon} of Appendix) and arguably contain richer language than \sst (see examples in Table \ref{table:dataset_example} of Appendix).

Table \ref{table:suda} presents the performance of \st and \tapt in \stl setting.
We find that domain shifts in unlabelled data consistently hurt the performance of \st, depending on the similarity between the source and target domains. 
The performance of \st drops sharply if the source and target domains are vastly different. For example, when \sst is used as the labelled data (target domain) and \imdb or \amazon is used as unlabelled data (source domain), the performance of \st falls from over 80\% to around 60\% or lower.
On the other hand, when using \sst and \imdb as the source and target domains, the performance of \st drops by a much smaller margin (a few percentage points). This shows the importance of training \st approaches using more informative labelled data, which is also consistent with our findings in \sect{paragraph:52}.

\tapt in the \stl setting is in fact a variation of \textit{domain adaptive pre-training} \cite{beltagy-etal-2019-scibert,gururangan-etal-2020-dont} applied to \ssl tasks.
Table \ref{table:suda} shows that the performance of \tapt remains stable when there exist domain shifts in the unlabelled data. 
Using more informative unlabelled data can further improve the performance of \tapt. For example, using \imdb or \amazon as unlabelled data when \sst is a target task, we see an improvement of about 4\% with 100 labelled samples. 
However, it is worth noting that \st methods can still be competitive compared to \tapt if the source and target domains are relatively similar. For instance, when using \amazon and \imdb as the source and target domains, \st still achieves better results than \tapt. 

%% file: paper/8_related_work.tex
\paragraph{Leveraging unlabelled data by Continuing Pre-training.} 
Previous work has shown that further pre-training LMs on the unlabelled data of a task \cite[e.g.][]{alsentzer-etal-2019-publicly,Mehri2020DialoGLUEAN,margatina-etal-2022-importance} or in-domain data \cite[e.g.][]{logeswaran-etal-2019-zero,gururangan-etal-2020-dont,xue-etal-2021-mt5} is beneficial to downstream tasks. However, it is unknown whether this is valid in \ssl settings. 
Previous studies in computer vision \cite{10.5555/3495724.3496047} and speech recognition \cite{xu2021self} have compared PT and \st.
However, our study has a different focus, specifically, we compare \tapt and \st in \nlp tasks.
Concurrently to our work, \citet{shi2023don} put forward prompt-based continued pre-training, which primarily aims to enhance the performance of prompt-based fine-tuning techniques \cite{schick-schutze-2021-exploiting,gao-etal-2021-making}. This approach outperforms these state-of-the-art \st approaches \cite{sohn2020fixmatch,xu2021dash,zhang2021flexmatch,berthelot2021adamatch} as well as the conventional \texttt{CLS}-based fine-tuning with \tapt.

\paragraph{Semi-supervised Learning.}
Recent work in \ssl has demonstrated great progress in effectively exploiting unlabelled data. A wide range of approaches has been proposed including Pseudo Labeling \cite{lee2013pseudo}, Temporal Ensemble \cite{DBLP:conf/iclr/LaineA17}, Mean Teacher~\cite{tarvainen2017mean}, Virtual Adversarial Training \cite{miyato2018virtual}, FixMatch~\cite{sohn2020fixmatch}.
A major issue for \st approaches is \textit{confirmation bias}, where the student model would accumulate errors from the teacher model when learning with inaccurate pseudo-labels \cite[e.g.][]{wang2021selftuning,goel2022pars,DST2022}.

While many efforts towards \st \cite[e.g.][]{ruder_strong_2018,gururangan-etal-2019-variational,li_semi-supervised_2019,chen-etal-2020-mixtext,meng-etal-2020-text,chen-etal-2020-local,He2020Revisiting,gera2022zero} have been made in \nlp, the performance of \st approaches across various labelled and unlabelled sizes has yet to be thoroughly explored. Although \citet{NEURIPS2020_f23d125d,li-etal-2021-task-adaptive} noted that training \st approaches from \tapt checkpoints can improve the performance, the performance of \tapt in \ssl tasks has not been either well-researched by previous works or compared with state-of-the-art \st approaches.

%% file: paper/9_conclusion.tex
In this work, we shed light on how \tapt performs against state-of-the-art \st approaches in various \ssl settings. Our experiments reveal that  \tapt achieves strong and robust performance, even with just a few hundred unlabelled examples. We further demonstrate that the \st approaches are vulnerable to small amounts of either labelled or unlabelled data.
We also find that \tapt is more robust than \st approaches in joint domain adaptation and \ssl settings. Overall, our empirical study demonstrates that \tapt is a  strong \ssl learner, competitive to more sophisticated \st approaches. In future work, we plan to further explore the potential of \tapt with unsupervised learning signals.

%% file: paper/limitations.tex
For easier comparison with previous work, we only focus on text classification tasks, while \st can also be applied to a variety of \nlp tasks, such as language generation, conversational systems and commonsense reasoning \cite{kedzie-mckeown-2019-good,He2020Revisiting,shi-etal-2022-learning,stepGame2022shi,10.1007/978-3-030-99736-6_20}.
We also assume that the datasets are roughly balanced. However, real-world datasets are usually class-imbalanced \cite{10.5555/2283696.2283708}, which might impact the performance of \tapt and \st. 
While this is out of the scope of this paper, we believe that this is an interesting avenue for future work. 
Additionally, different labelled and unlabelled sizes may impact the performance of \st approaches in the domain shift setting.
However, this doesn't alter our conclusion that the effectiveness of \st approaches significantly fluctuates across different scenarios.

%% file: paper/appendix.tex
\clearpage
\appendix
\section*{Appendix Overview}
The appendix is structured as follows:
\paragraph{Appendix \sect{sec:dataset}} provides a brief description and example for each dataset (subsection \sect{appendix:description}). Additionally, a similarity analysis among datasets and an illustration of overlaps between \imdb and \amazon are included (subsection \sect{appendix:dataset_similarity}).

\paragraph{Appendix \sect{appendix:baseline_models}} presents a brief description of state-of-the-art \st approaches.

\paragraph{Appendix \sect{sec:falling_below_baseline}} includes a supplementary Table that examines the effect of low unlabelled data sizes.

\paragraph{Appendix \sect{sec:verification}} presents additional experiments to verify our findings using other \st approaches.

\paragraph{Appendix \sect{sec:justification}} includes additional experiments to train \st approaches using \tapt checkpoints.

\paragraph{Appendix \sect{sec:training_details}} provides implementation details and hyperparameters for \tapt, \st, and \ft methods used in our experiments.

\section{Datasets}
In this section, we briefly introduce the datasets used in our work and provide additional analysis of the similarity among them. Specifically, we provide four examples to demonstrate the overlap between \imdb and \amazon, as a supplement to our domain adaptation analysis (\sect{sec:da}).

\label{sec:dataset}
\subsection{Description}
\label{appendix:description}
In this section, we briefly introduce \imdb, \sst, \ag, \amazon, %\yelp, 
and \yahoo datasets. Table \ref{table:dataset_example} list examples for each dataset.

\paragraph{\imdb.} 
The \imdb dataset \cite{maas-etal-2011-learning} contains a collection of $50\,000$ reviews from the Internet Movie Database, with no more than 30 reviews per movie.
This dataset contains an equal number of positive and negative reviews, yielding a 33\% Marco-$F_1$ score for random guessing. There are $25\,000$ and $25\,000$ for training and testing, respectively. We follow \citet{wang2022usb} to split the dataset by selecting $12\,500$ samples and $1\,000$ samples per class from the train set to form a train and validation set, respectively.

\paragraph{\sst.} The \sst dataset \cite{wang-etal-2018-glue} consists of sentences from movie reviews and human annotations of their sentiment. The task is to predict the sentiment of a given sentence. Similar to \imdb, this is also a binary classification task. There are $67\,349$ and $872$ for training and testing. We select $60\,000$ and $7\,349$  samples from the train set to form a train and validation set, respectively, where the validation set contains $3\,675$ and $3\,674$ samples for two classes, respectively.

\paragraph{\ag.}
The \ag topic classification dataset is constructed by \citet{10.5555/2969239.2969312}, where 4 classes are used. Each class contains $30\,000$ training samples and $1\,900$ test samples. We follow \citet{wang2022usb} to split the dataset by selecting $25\,000$ samples and $2\,500$ samples per class from the train set samples to form a train and validation set, respectively.

\paragraph{\amazon.}
The \amazon dataset \cite{10.1145/2507157.2507163} is a sentiment classification dataset, with five classes. There are $600\,000$ train samples and $130\,000$ test samples per class.  We follow \citet{wang2022usb} to split the dataset by selecting $50\,000$ samples and $5\,000$ samples per class from the train set samples to form a train and validation set, respectively.

\paragraph{\yahoo.}
The \yahoo dataset \cite{chang2008importance} is a topic classification dataset, with ten classes. There are $140\,000$ train samples and $6\,000$ test samples per class.  We follow \citet{wang2022usb} to split the dataset by selecting $50\,000$ samples and $5\,000$ samples per class from the train set samples to form a train and validation set, respectively.

\begin{figure*}[t!]
  \centering
  \includegraphics[width=0.6\textwidth]{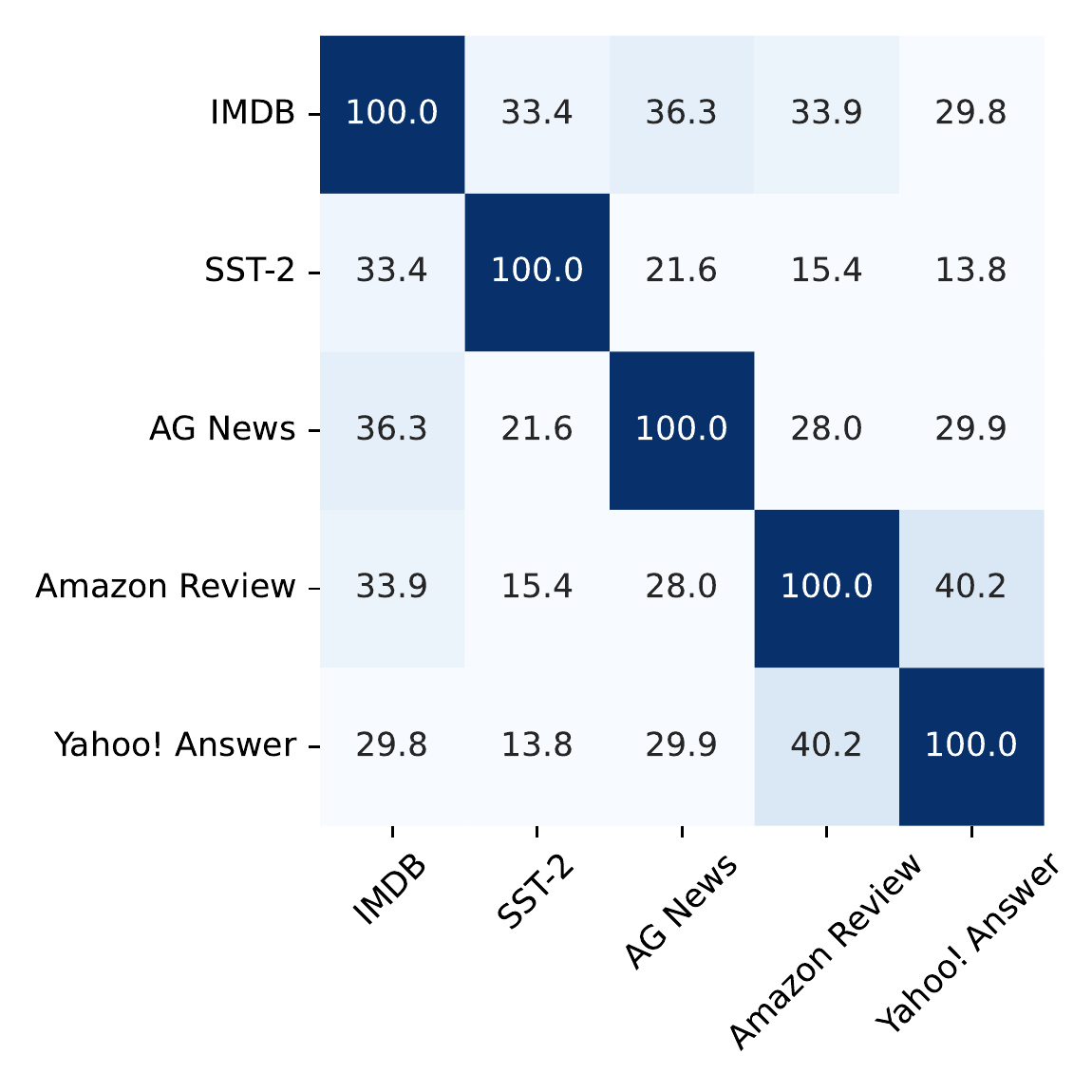}
  \caption{Vocabulary overlap (\%) across datasets.}
  \label{fig:overlap}
\end{figure*}

\subsection{Dataset Similarity}
\label{appendix:dataset_similarity}
We provide an analysis of the vocabulary overlap of the datasets, as shown in Figure \ref{fig:overlap}. Additionally, in Table \ref{table:overlap_imdb_and_amazon}, we provide some examples to illustrate the overlap between \imdb and \amazon.

As shown in Table \ref{table:dataset_example}, although both the \sst and \imdb datasets are sentiment analysis tasks for movie reviews, the \sst datasets contain shorter and vaguer sentences than the \imdb dataset. This difference could be a potential reason for poor performance of \st approaches in the \uda setting (\sect{paragraph:uda}).
In contrast, the \amazon dataset, which is a product review sentiment analysis dataset, is more similar to the \imdb dataset than the \sst dataset, as shown in Table \ref{table:overlap_imdb_and_amazon}.
This suggests a potential reason for the performance of \st and \tapt in the \stl setting (\sect{paragraph:stl}).

\section{\st Frameworks}
\label{appendix:baseline_models}
\paragraph{\vat.} \vat \cite{miyato2018virtual} proposed a regularization technique that forces pairs of data points that are very close in the input space to be close to each other in the output space. \vat adds small perturbation to the input data and forces the model to produce similar predictions.

\paragraph{\fixmatch.} \fixmatch \cite{sohn2020fixmatch} generates artificial labels using both consistency regularization and pseudo-labelling, where the artificial labels are produced based on weakly-augmented unlabelled data. These artificial labels are then used as targets to train the model on strongly-augmented unlabelled data. \fixmatch only retains an artificial label if the model assigns a high probability to one of the possible classes.

\paragraph{\dash.} \dash \cite{xu2021dash} extends \fixmatch by introducing a mechanism with a dynamically adjusted threshold of loss to select a subset of training examples from the unlabelled data for performing \ssl.

\paragraph{\flexmatch.} \flexmatch \cite{zhang2021flexmatch} also extends \fixmatch by introducing the concept of curriculum learning \cite{bengio2009curriculum} to flexibly adjust thresholds for different classes at each time step and select unlabelled data and their pseudo labels that are more likely to be informative.

\paragraph{\adamatch.} \adamatch \cite{berthelot2021adamatch} aims to solve domain adaptation problems in \ssl and build a high-accuracy model that trains on and tests on different data distributions. \adamatch builds on \fixmatch and introduces a relative confidence threshold and a modified distribution alignment from \cite{berthelot2019remixmatch}.

\section{Probability of performing worsen than \partials.} 
\label{sec:falling_below_baseline}
In \sect{paragraph:unlabel_initialize}, we discuss that we select the model performance with the three lowest unlabelled sizes (the first three columns in Figure \ref{fig:heatmap}) for each dataset and exclude the model performance with the lowest labelled size (the last row in Figure \ref{fig:heatmap}). This results in 9 cells in \imdb, 3 cells in \sst, 9 cells in \amazon, and 12 cells in \yahoo, where \tapt has one run per cell and \st (\flexmatch and \adamatch) has two runs per cell. We consider a run to be a failure if its performance is worse than its corresponding \partials baseline.

Table \ref{table:falling_below_baseline} lists the probability of \st and \tapt of falling below the \partials baseline with selected combinations of labelled and unlabelled sizes. 

\section{Further validations with other \st approaches}
\label{sec:verification}
In this section, we conduct additional experiments on \st approaches, including \vat, \dash, and \fixmatch to demonstrate that our findings are applicable to other \st approaches as well.

In Table \ref{table:verify_conclusion}, we select several combinations of labelled and unlabelled sizes on \imdb, \sst, \amazon, and \yahoo datasets. Our experimental results show that other \st approaches do not perform well when the labelled size is low, and that other \st approaches have a high probability to perform worsen than \partials baselines when the unlabelled size is low.
This suggests that poor performance when the labelled or unlabelled size is inadequate may be a common problem of state-of-the-art \st approaches.

\section{Train \st approaches with \tapt checkpoints}
\label{sec:justification}
Previous works \cite{NEURIPS2020_f23d125d,li-etal-2021-task-adaptive} have suggested that training \st approaches from a \tapt checkpoint may be beneficial. 
Here we also provide some additional experiments to train \st approaches with \tapt checkpoints to further corroborate our findings.

Table \ref{table:st+tapt_2} shows that \tapt outperforms \adamatch+\tapt or \flexmatch+\tapt with two different labelled sizes on the \yahoo dataset.

Table \ref{table:st+tapt} shows that training \st approaches from \tapt checkpoints could improve the performance of \st but cannot solve the issue of \st approaches when labelled or unlabelled data is not adequate. Specifically, the performance of \st+\tapt is still poor when labelled data is insufficient, as discussed in \sect{paragraph:52}. 
Meanwhile, in Table \ref{table:st+tapt}, the performance of \st+\tapt could be outperformed by the \partials baselines when unlabelled data is inadequate, while \tapt consistently outperforms the \partials baselines. 
When the labelled size is 10, the performance of \st trained with fewer unlabelled samples tends to be better, indicating that reducing the number of unlabelled data can be helpful, as discussed in \sect{paragraph:case_3}.

\section{Implementation Details}
\label{sec:training_details}
We consistently use five random seeds, ranging from 1 to 5, for all algorithms. The sampled labelled data is the same for all algorithms for a given seed. The development and test sets remain unchanged for all different labelled and unlabelled data sizes.

Our model implementation uses open-source libraries including HuggingFace Transformers\footnote{https://huggingface.co}, Fairseq\footnote{https://github.com/facebookresearch/fairseq}, and USB\footnote{https://github.com/microsoft/Semi-supervised-learning}. Our experiments of \tapt are performed on 8x32GB V100 GPUs, with a batch size of 16 per device and 2 gradient accumulation steps.

Table \ref{table:tapt_hyperparameters} lists the hyperparameters used for the \tapt phrase. 
Table \ref{table:finetune_hyperparameters} lists the hyperparameters used for the fine-tuning phrase. 
Table \ref{table:semi_hyperparameters} lists the hyperparameters used for \st approaches.

\input{table/dataset_example}

\input{table/overlap_imdb_and_amazon}
\input{table/falling_below_baseline}
\input{table/verify_with_other_sts}
\input{table/st+tapt_2}
\input{table/st+tapt}
\input{table/tapt_hyparameters}

%% file: table/dataset_example.tex
\begin{table*}[ht!]
\centering
\small
\caption{Examples for datasets.}
\label{table:dataset_example}
\begin{adjustbox}{max width=\textwidth}
\begin{tabular}{>{\raggedright}p{3cm}p{15cm}}
\toprule
\bf Dataset     & \bf Example  \\ 
\midrule
\imdb           & I watched this movie after seeing other comments on IMDb, even convincing my wife that it was a "unique horror movie." I wanted to like this movie, but was unable to.The "love story" was good, but the horror aspect was quite bad. If the story was just about a young man who fell in love with a girl suffering from parasomnia, then it would have been a better movie.The care centre stretched credulity well past the limits, in fact it was quite ridiculous. The doctor happily ignors privacy laws and professionalism. A nurse goes into a room for a routine feeding of a dangerous patient (without security escort), and drops the tray and runs out of the room screaming for no apparent reason. The forensic patient (and the film's villain) is tied up in a standing position fully clothed - apparently for years? None of it makes much sense.The movie even had some actors that I've liked in other things, such as the detectives, but still I can't recommend this movie. \\
\midrule
\sst            & a rewarding work of art for only the most patient and challenge-hungry moviegoers. \\
\midrule
\ag             & Teen flies in plane \#39;s landing gearA homeless teenager who hid in the landing gear of a passenger plane survived a 700-kilometre flight across south-western China but his companion fell and probably died, state media reported on Friday. \\
\midrule
\amazon         & THIS is MUSIC at its BESTRob Dougan has done it. He's crafted musical perfection, or close to it anyway. I have finally found the music I've been waiting for my whole life in this album - Rob D you are a genius. I think a lot of us wanted to know more about this guy as soon as we heard the track playing to the "Woman in the Red Dress" scene. Now I know why the Wachowski brothers have enlisted his musical talents to flesh out their movies.I know I should be trying to write a more helpful, objective review but I can do nothing but wax poetic for Rob Dougan and his debut album. He has mixed classical melodies with awesome electric beats and it all comes together in an audio orgy. Just buy the album already and let's get Rob some more mainstream recognition. \\
\midrule
\yahoo          & Does anybody know a great deal about angels? I'm looking for names, if they're good or bad, what they look like, etc.  The more detail the better.  All religions accepted \\
\bottomrule
\end{tabular}
\end{adjustbox}
\end{table*}

%% file: table/overlap_imdb_and_amazon.tex
\begin{table*}
\centering
\small
\caption{Similarity analysis between \imdb and \amazon with four examples that highlight the overlap.}
\resizebox{\textwidth}{!}{
\begin{tabular}{>{\raggedright}p{8cm}p{8cm}}
\toprule 
\textbf{\imdb}  & \textbf{\amazon}\\
\midrule 
I loved this \textbf{movie} since I was 7 and I saw it on the opening day. It was so \textbf{touching} and beautiful. I strongly recommend seeing for all. It's a \textbf{movie} to watch with your family by far. My MPAA rating: PG-13 for thematic elements, prolonged scenes of disastor, nudity/sexuality and some language. 
& 
This is a very \textbf{touching}, spiritual \textbf{movie}! When I first saw this film, [...]. I was deeply moved by this motion picture, and the DVD brings the story to your own home. The bonus materials could be better, but the main part of the DVD is the actual \textbf{movie}. Great, great, great film... [...] \\
\midrule
Pacino is over-the-top but to good effect as he's clearly having loads of \textbf{fun}. Beatty is \textbf{great} [...] The lighting, velvet overtones and smog/smoke combine to create a \textbf{great} effect.There are some really \textbf{funny} cameos [...] \textbf{Highly recommended}. 4.5/5 stars. [...] 
& 
Makes a \textbf{great} gift! We bought this book for my dad for Father's Day this year, and thought he would have \textbf{fun} reading it since he has four granddaughters. He loved it and has even selected stories to read to the girls during over-nights with Grandpa and Grandma. I \textbf{highly recommend} it as a \textbf{great} gift. \\
\midrule
The late [...] scripted this tale of \textbf{terror} and it was absolutely one of the \textbf{scariest movies} I ever saw as a kid. (I had to walk MILES just to see a \textbf{movie}, and it was usually dark when I emerged from the theater; seeing a horror \textbf{movie} was always unnerving [...] & Movia ... please .... This \textbf{movie} is a masterpiece of \textbf{terror} \& suspence \& Beautifully filmed \& acted.Comparisons to reality are not allowed when reviewing films of this caliber. Your reaction (though it MAY be \textbf{sarcastic}) is EXACT proof of it's genius! Watch it again...and this time....bask in all it's glory! \\
\midrule
Fabulous actors, beautiful scenery, stark reality [...] I tried to buy the video for several years, finally bought it used from a video store that went out of business. But Yippee! The DVD is now for sale, I purchased it on amazon.com. Not cheap, but \textbf{well worth} it to me. [...]
& 
\textbf{Well worth} the import price. My first impression of this album was a good one, but as time went on it came to grow on me more and more. This is certainly one of the better Costes albums. The mixing is nothing revolutionary, but it is well done and all tracks flow into each other very well. [...]. \\
\bottomrule
\end{tabular}}
\label{table:overlap_imdb_and_amazon}
\end{table*}

%% file: table/falling_below_baseline.tex
\begin{table*}[ht!]
\centering
% \footnotesize
% \small
\caption{Results on the effect of low unlabelled sizes on \st and \tapt. Failure means performing worsen than \partials.}
\label{table:falling_below_baseline}
\begin{adjustbox}{max width=\textwidth}
\begin{tabular}{lllrr}
\toprule
\bf Task       & \bf \#Unl.          & \bf \#Lab.                       & \bf Prob. of \st Failure  & \bf Prob. of \tapt Failure \\ 
\midrule
\imdb          & 100, 500, 2k        & 10, 20, 200, 1k                  &  6/18 (33\%)    &  0/9 (0\%)   \\
\sst           & 100, 500, 2k        & 40, 200, 1k, 5k                  &  4/6 (67\%)      &  0/3 (0\%)    \\
\amazon        & 1k,  5k, 20k        & 100, 500, 2k, 10k                &  10/18 (56\%)    &  0/9 (0\%)   \\
\yahoo         & 1k,  5k, 20k        & 20, 100, 500, 2k, 10k            &  13/24 (54\%)    &  0/12 (0\%)   \\
\bottomrule
\end{tabular}
\end{adjustbox}
\end{table*}

%% file: table/verify_with_other_sts.tex
\begin{table*}[hbt!]
\centering
% \scriptsize
% \setlength\tabcolsep{2pt}
\caption{We further verify our conclusion on \vat, \dash, \fixmatch that . We report the average Macro-$F_1$ score on the test set across five seeds, with standard deviations as subscripts. Blue represents the best results for each row.}
\label{table:verify_conclusion}
\resizebox{\textwidth}{!}{
\begin{tabular}{lccccccccc}
\toprule
\bf Dataset          & \bf \#Unl.       & \bf \#Lab.  & \bf \vat                & \bf \fixmatch         & \bf \dash       &  \bf \flexmatch         & \bf \adamatch         & \bf \tapt    & \bf \partials\\
\midrule
\multirow{7}{*}{\imdb}    & 100              & 4           & 33.5$_{0.2}$            & 33.4$_{0.1}$          & 33.4$_{0.1}$    & 35.7$_{4.2}$            & 34.1$_{0.7}$          & \hl61.8$_{6.7}$  & 59.4$_{4.8}$       \\
                          & 100              & 10          & 61.6$_{20.1}$            & 45.4$_{21.6}$         & 34.7$_{2.2}$    & 49.0$_{19.9}$           & 52.4$_{21.0}$         & \hl75.5$_{6.9}$  & 71.8$_{8.5}$     \\
                          & 100              & 20          & \hl87.1$_{2.2}$            & 64.6$_{16.5}$         & 67.8$_{16.6}$   & 85.5$_{2.9}$            & 79.1$_{7.6}$          & 85.5$_{1.0}$  & 84.1$_{1.9}$      \\
                          & 500              & 4           & 33.4$_{0.0}$            & 33.4$_{0.1}$          & 33.4$_{0.1}$    & 33.4$_{0.1}$            & 33.6$_{0.3}$          & \hl63.4$_{7.2}$  & 58.2$_{7.1}$     \\
                          & 2k               & 4           & 33.3$_{0.0}$            & 33.3$_{0.0}$          & 33.3$_{0.0}$    & 33.3$_{0.0}$            & 33.3$_{0.0}$          & \hl63.1$_{6.2}$ & 60.9$_{5.6}$     \\
                          & 10k              & 4           & 33.3$_{0.0}$            & 33.5$_{0.3}$          & 33.3$_{0.0}$    & 34.0$_{1.2}$            & 33.6$_{0.4}$          & \hl64.1$_{8.9}$  & 62.4$_{7.9}$     \\
                          & 23k              & 4           & 33.3$_{0.0}$            & 33.3$_{0.0}$          & 57.4$_{29.4}$   & 45.3$_{23.9}$           & 33.3$_{0.0}$          & \hl68.8$_{5.6}$  & 65.6$_{10.4}$     \\
\midrule
\multirow{6}{*}{\sst}     & 100              & 40          & 63.3$_{10.6}$           & 46.9$_{9.7}$          & 47.9$_{7.0}$    & 57.2$_{4.5}$            & 51.0$_{14.0}$         & \hl78.7$_{2.5}$  & 76.4$_{3.7}$  \\
                          & 500              & 40          & 55.7$_{16.8}$           & 53.8$_{8.9}$          & 51.2$_{10.0}$   & 67.7$_{10.7}$           & 59.1$_{11.4}$         & \hl83.3$_{4.8}$  & 72.9$_{7.9}$    \\
                          & 500              & 200         & 83.0$_{1.6}$            & 84.5$_{2.8}$          & 82.6$_{3.5}$    & 83.8$_{3.0}$            & 87.4$_{1.9}$          & \hl88.8$_{0.9}$  & 88.3$_{0.9}$     \\
                          & 2k               & 40          & 55.9$_{24.2}$           & 36.4$_{3.0}$          & 35.3$_{2.0}$    & 56.6$_{6.7}$            & 49.3$_{13.8}$         & \hl79.3$_{5.9}$  & 71.7$_{8.2}$    \\
                          & 10k              & 40          & 73.5$_{20.5}$           & 38.9$_{11.4}$         & 35.6$_{2.6}$    & 56.9$_{12.5}$           & 36.2$_{2.9}$          & \hl85.9$_{1.0}$  & 78.5$_{7.5}$    \\
                          & 60k              & 40          & 79.6$_{13.4}$           & 32.6$_{1.7}$          & 33.4$_{0.6}$    & 40.6$_{7.7}$            & 42.6$_{13.3}$         & \hl82.6$_{4.0}$  & 75.3$_{7.2}$    \\
\midrule
\multirow{7}{*}{\amazon}  & 1k               & 20          & 13.5$_{5.2}$            & 14.9$_{5.6}$          & 20.3$_{3.0}$    & 25.8$_{3.2}$            & 20.7$_{1.1}$          & 32.0$_{1.8}$  & \hl32.5$_{2.2}$  \\
                          & 1k               & 100         & 46.1$_{2.2}$            & 36.3$_{3.1}$          & 35.3$_{6.2}$    & 43.4$_{1.7}$            & 40.3$_{2.2}$          & \hl48.5$_{0.9}$  & 48.2$_{2.2}$      \\
                          & 1k               & 500         & 52.6$_{0.2}$            & 50.8$_{1.5}$          & 49.5$_{1.0}$    & 54.1$_{1.0}$            & 52.8$_{1.1}$          & \hl55.9$_{0.3}$  & 55.3$_{0.5}$      \\
                          & 5k               & 20          & 15.5$_{7.8}$            & 13.5$_{3.3}$          & 22.2$_{5.2}$    & 23.2$_{7.3}$            & 16.9$_{6.9}$          & \hl32.8$_{3.4}$  & 32.3$_{2.5}$     \\
                          & 20k              & 20          & 19.3$_{7.5}$            & 15.2$_{3.9}$          & 20.5$_{6.4}$    & 19.1$_{10.0}$           & 19.3$_{6.3}$          & \hl32.0$_{3.2}$  & 31.6$_{3.6}$     \\
                          & 100k             & 20          & 14.1$_{7.3}$            & 11.9$_{2.9}$          & 20.7$_{5.2}$    & 15.3$_{2.6}$            & 12.5$_{3.7}$          & 30.7$_{3.6}$  & \hl30.8$_{3.9}$    \\
                          & 250k             & 20          & 10.3$_{5.0}$            & 10.9$_{3.6}$          & 22.0$_{5.7}$    & 22.7$_{4.9}$            & 14.4$_{5.6}$          & 30.2$_{2.4}$  & \hl32.1$_{3.1}$    \\ % Checking tapt pooh
\midrule
\multirow{8}{*}{\yahoo}   & 1k               & 10          & 1.9$_{0.1}$            & 2.0$_{0.1}$           & 4.6$_{2.9}$      & 15.7$_{2.6}$            & 18.8$_{7.9}$          & \hl29.6$_{5.8}$    & 23.5$_{4.5}$     \\
                          & 1k               & 20          & 6.7$_{2.8}$            & 10.1$_{4.2}$          & 9.6$_{3.2}$      & 32.7$_{9.1}$            & 28.8$_{5.8}$          & \hl38.9$_{4.1}$    & 34.1$_{3.6}$   \\
                          & 1k               & 100         & 55.2$_{1.7}$           & 46.9$_{4.4}$          & 45.3$_{3.7}$     & 54.2$_{1.4}$            & 53.9$_{1.3}$          & \hl59.7$_{0.8}$    & 57.4$_{1.6}$  \\
                          & 1k               & 500         & 59.2$_{0.4}$           & 61.6$_{0.6}$          & 60.7$_{1.3}$     & 61.9$_{1.1}$            & 61.5$_{0.9}$          & \hl65.8$_{0.3}$    & 65.5$_{0.2}$ \\
                          & 5k               & 10          & 1.8$_{0.0}$            & 3.2$_{2.6}$           & 3.7$_{2.7}$      & 16.4$_{10.8}$           & 17.8$_{11.7}$         & \hl31.4$_{5.1}$    & 25.7$_{3.9}$    \\
                          & 20k              & 10          & 2.4$_{0.9}$            & 2.0$_{0.3}$           & 4.9$_{3.1}$      & 7.3$_{4.7}$             & 25.2$_{12.2}$         & \hl32.4$_{5.6}$    & 27.2$_{4.4}$  \\
                          & 100k             & 10          & 2.3$_{0.6}$            & 3.8$_{2.5}$           & 3.4$_{2.9}$      & 2.9$_{1.1}$             & 17.7$_{11.4}$         & \hl30.8$_{3.8}$    & 28.0$_{5.0}$     \\
                          & 500k             & 10          & 2.0$_{0.4}$            & 1.8$_{0.0}$           & 2.6$_{1.2}$      & 2.5$_{0.9}$             & 14.3$_{6.0}$          & \hl27.3$_{4.6}$    & 24.7$_{4.8}$    \\
\bottomrule
\end{tabular}
}
\end{table*}

%% file: table/st+tapt_2.tex
\begin{table*}[!ht]
\centering
\footnotesize
\caption{Results of \adamatch+\tapt and \flexmatch+\tapt on \yahoo with two different labelled sizes.}
\label{table:st+tapt_2}
\begin{adjustbox}{max width=\columnwidth}
\begin{tabular}{lcc}
\toprule
                   & \multicolumn{2}{c}{\bf \yahoo} \\ 
                   \cmidrule{2-3}       
                      &  500              &  2000         \\ 
\midrule
\adamatch          & 68.0$_{0.7}$      & 69.5$_{0.3}$     \\
\quad + \tapt      & 68.2$_{1.0}$      & 69.8$_{0.3}$     \\
\midrule
\flexmatch          & 66.6$_{0.7}$      & 68.7$_{0.4}$    \\
\quad + \tapt       & 66.7$_{1.2}$      & 69.0$_{0.5}$     \\
\midrule
\partials           & 65.4$_{0.3}$      & 68.5$_{0.3}$ \\
\quad + \tapt      & 68.8$_{0.7}$      & 71.5$_{0.3}$ \\
\midrule
\fullys.         & \multicolumn{2}{c}{75.3$_{0.2}$} \\
\quad + \tapt    & \multicolumn{2}{c}{75.4$_{0.1}$} \\
\bottomrule
\end{tabular}
\end{adjustbox}
\end{table*}

%% file: table/st+tapt.tex
\begin{table*}[hbt!]
\centering
\scriptsize
\caption{We further verify our conclusion on \flexmatch+\tapt. We report the average Macro-$F_1$ score on the test set across five seeds, with standard deviations as subscripts. Blue represents the best results for each row.}
\label{table:st+tapt}
\resizebox{\textwidth}{!}{
\begin{tabular}{lcccccc}
\toprule
\bf Dataset          & \bf \#Unl.       & \bf \#Lab.  & \bf \flexmatch + \tapt    &  \bf \flexmatch       & \bf \tapt          & \bf \partials\\
\midrule
\multirow{8}{*}{\yahoo}   & 1k               & 10          & 17.0$_{4.9}$              & 15.7$_{2.6}$          & \hl29.6$_{5.8}$    & 23.5$_{4.5}$     \\
                          & 1k               & 20          & \hl39.4$_{2.0}$           & 32.7$_{9.1}$          & 38.9$_{4.1}$       & 34.1$_{3.6}$   \\
                          & 1k               & 100         & 55.2$_{1.8}$              & 54.2$_{1.4}$          & \hl59.7$_{0.8}$    & 57.4$_{1.6}$  \\
                          & 1k               & 500         & 62.0$_{0.7}$              & 61.9$_{1.1}$          & \hl65.8$_{0.3}$    & 65.5$_{0.2}$ \\
                        %   & 5k               & 10          & \hl35.1$_{7.7}$           & 16.4$_{10.8}$         & 31.4$_{5.1}$       & 25.7$_{3.9}$    \\
                          & 20k              & 10          & 4.0$_{1.4}$               & 7.3$_{4.7}$           & \hl32.4$_{5.6}$    & 27.2$_{4.4}$  \\
                          & 100k             & 10          & 5.1$_{6.1}$               & 2.9$_{1.1}$           & \hl30.8$_{3.8}$    & 28.0$_{5.0}$     \\
                          & 500k             & 10          & 2.5$_{1.1}$               & 2.5$_{0.9}$           & \hl27.3$_{4.6}$    & 24.7$_{4.8}$    \\
\bottomrule
\end{tabular}
}
\end{table*}

%% file: table/tapt_hyparameters.tex
% \clearpage
\begin{table*}[t!]
    \centering
    \small
    \begin{tabular}{cc}
        \toprule
        \textbf{Hyperparameter} & \textbf{Assignment}  \\
        \midrule
        number of steps & 100 epochs\\
        \midrule
        batch size & 256 \\
        \midrule
        maximum learning rate & 1e-06, 1e-4 \\
        \midrule
        learning rate optimizer & AdamW \\
        \midrule
        Adam epsilon & 1e-6 \\
        \midrule
        Adam beta weights & 0.9, 0.98\\
        \midrule
        learning rate scheduler & Warmup linear \\
        \midrule
        Weight decay & 0.01 \\
        \midrule
        Warmup proportion & 0.06 \\
        \midrule
        learning rate decay & linear \\
        \bottomrule
    \end{tabular}
    \caption{Hyperparameters for task-adaptive pretraining. The learning rate and unlabelled size are tightly connected and need to be adjusted together. We generally recommend increasing the learning rate as you increase the unlabelled size. Different from its predecessor, \textsc{Bert} \cite{devlin2018bert}, where the next sentence prediction objective is used, \roberta \cite{liu2019roberta} is only trained with the \mlm objective (i.e., cross-entropy loss on predicting randomly masked tokens), dynamically changing the masking pattern applied to the training examples and typically using the masking probability of 0.15.
    } 
    \label{table:tapt_hyperparameters}
\end{table*}

\begin{table*}[t!]
    \centering
    \small
    \begin{tabular}{cc}
        \toprule
        \textbf{Hyperparameter} & \textbf{Assignment}  \\
        \midrule
        number of steps & 10 or 50 epochs \\
        \midrule
        batch size & 16 or 32 \\
        \midrule
        maximum learning rate & 2e-05 \\
        \midrule
        learning rate optimizer & AdamW \\
        \midrule
        maximum sequence length & 256 \\
        % \midrule
        % Adam beta weights & 0.9, 0.98\\
        \midrule
        learning rate scheduler & Warmup linear \\
        % \midrule
        % Weight decay & 0.01 \\
        \midrule
        Warmup proportion & 0.06 \\
        \midrule
        learning rate decay & linear \\
        \bottomrule
    \end{tabular}
    \caption{Hyperparameters for fine-tuning. More epochs are used when the labelled size is low.} 
    \label{table:finetune_hyperparameters}
\end{table*}

\begin{table*}[t!]
    \centering
    \small
    \begin{tabular}{cc}
        \toprule
        \textbf{Hyperparameter} & \textbf{Assignment}  \\
        \midrule
        number of steps & $25\,600$ or $51\,200$ steps \\
        \midrule
        batch size & 16 \\
        \midrule
        maximum learning rate & 2e-05 \\
        \midrule
        learning rate optimizer & AdamW \\
        \midrule
        maximum sequence length & 256 \\
        % \midrule
        % Adam beta weights & 0.9, 0.98\\
        \midrule
        learning rate scheduler & Warmup linear \\
        % \midrule
        % Weight decay & 0.01 \\
        \midrule
        Warmup proportion & 0.05 \\
        \midrule
        learning rate decay & linear \\
        \bottomrule
    \end{tabular}
    \caption{Hyperparameters for self training. Algorithm-specific hyperparameters will be released in configuration files with the code.} 
    \label{table:semi_hyperparameters}
\end{table*}

%% file: acl2023.bbl
\begin{thebibliography}{72}
\expandafter\ifx\csname natexlab\endcsname\relax\def\natexlab#1{#1}\fi

\bibitem[{Alsentzer et~al.(2019)Alsentzer, Murphy, Boag, Weng, Jindi, Naumann,
  and McDermott}]{alsentzer-etal-2019-publicly}
Emily Alsentzer, John Murphy, William Boag, Wei-Hung Weng, Di~Jindi, Tristan
  Naumann, and Matthew McDermott. 2019.
\newblock \href {https://doi.org/10.18653/v1/W19-1909} {Publicly available
  clinical {BERT} embeddings}.
\newblock In \emph{Proceedings of the 2nd Clinical Natural Language Processing
  Workshop}, pages 72--78, Minneapolis, Minnesota, USA. Association for
  Computational Linguistics.

\bibitem[{Arazo et~al.(2020)Arazo, Ortego, Albert, O’Connor, and
  McGuinness}]{arazo2020pseudo}
Eric Arazo, Diego Ortego, Paul Albert, Noel~E O’Connor, and Kevin McGuinness.
  2020.
\newblock \href {https://arxiv.org/abs/1908.02983} {Pseudo-labeling and
  confirmation bias in deep semi-supervised learning}.
\newblock In \emph{2020 International Joint Conference on Neural Networks
  (IJCNN)}, pages 1--8. IEEE.

\bibitem[{Artetxe et~al.(2018)Artetxe, Labaka, and
  Agirre}]{artetxe-etal-2018-robust}
Mikel Artetxe, Gorka Labaka, and Eneko Agirre. 2018.
\newblock \href {https://doi.org/10.18653/v1/P18-1073} {A robust self-learning
  method for fully unsupervised cross-lingual mappings of word embeddings}.
\newblock In \emph{Proceedings of the 56th Annual Meeting of the Association
  for Computational Linguistics (Volume 1: Long Papers)}, pages 789--798,
  Melbourne, Australia. Association for Computational Linguistics.

\bibitem[{Beltagy et~al.(2019)Beltagy, Lo, and
  Cohan}]{beltagy-etal-2019-scibert}
Iz~Beltagy, Kyle Lo, and Arman Cohan. 2019.
\newblock \href {https://doi.org/10.18653/v1/D19-1371} {{S}ci{BERT}: A
  pretrained language model for scientific text}.
\newblock In \emph{Proceedings of the 2019 Conference on Empirical Methods in
  Natural Language Processing and the 9th International Joint Conference on
  Natural Language Processing (EMNLP-IJCNLP)}, pages 3615--3620, Hong Kong,
  China. Association for Computational Linguistics.

\bibitem[{Ben-David et~al.(2010)Ben-David, Blitzer, Crammer, Kulesza, Pereira,
  and Vaughan}]{ben2010theory}
Shai Ben-David, John Blitzer, Koby Crammer, Alex Kulesza, Fernando Pereira, and
  Jennifer~Wortman Vaughan. 2010.
\newblock \href {https://link.springer.com/article/10.1007/s10994-009-5152-4}
  {A theory of learning from different domains}.
\newblock \emph{Machine learning}, 79(1):151--175.

\bibitem[{Bengio et~al.(2009)Bengio, Louradour, Collobert, and
  Weston}]{bengio2009curriculum}
Yoshua Bengio, J\'{e}r\^{o}me Louradour, Ronan Collobert, and Jason Weston.
  2009.
\newblock \href {https://doi.org/10.1145/1553374.1553380} {Curriculum
  learning}.
\newblock In \emph{Proceedings of the 26th Annual International Conference on
  Machine Learning}, ICML '09, page 41–48, New York, NY, USA. Association for
  Computing Machinery.

\bibitem[{Berthelot et~al.(2019{\natexlab{a}})Berthelot, Carlini, Cubuk,
  Kurakin, Sohn, Zhang, and Raffel}]{berthelot2019remixmatch}
David Berthelot, Nicholas Carlini, Ekin~D Cubuk, Alex Kurakin, Kihyuk Sohn, Han
  Zhang, and Colin Raffel. 2019{\natexlab{a}}.
\newblock \href {https://arxiv.org/abs/1911.09785} {Remixmatch: Semi-supervised
  learning with distribution alignment and augmentation anchoring}.
\newblock \emph{arXiv preprint arXiv:1911.09785}.

\bibitem[{Berthelot et~al.(2019{\natexlab{b}})Berthelot, Carlini, Goodfellow,
  Papernot, Oliver, and Raffel}]{berthelot2019mixmatch}
David Berthelot, Nicholas Carlini, Ian~J. Goodfellow, Nicolas Papernot, Avital
  Oliver, and Colin Raffel. 2019{\natexlab{b}}.
\newblock \href
  {https://proceedings.neurips.cc/paper/2019/hash/1cd138d0499a68f4bb72bee04bbec2d7-Abstract.html}
  {Mixmatch: {A} holistic approach to semi-supervised learning}.
\newblock In \emph{Advances in Neural Information Processing Systems 32: Annual
  Conference on Neural Information Processing Systems 2019, NeurIPS 2019,
  December 8-14, 2019, Vancouver, BC, Canada}, pages 5050--5060.

\bibitem[{Berthelot et~al.(2022)Berthelot, Roelofs, Sohn, Carlini, and
  Kurakin}]{berthelot2021adamatch}
David Berthelot, Rebecca Roelofs, Kihyuk Sohn, Nicholas Carlini, and Alexey
  Kurakin. 2022.
\newblock \href {https://openreview.net/forum?id=Q5uh1Nvv5dm} {Adamatch: A
  unified approach to semi-supervised learning and domain adaptation}.
\newblock In \emph{International Conference on Learning Representations}.

\bibitem[{Cai and Lapata(2019)}]{cai-lapata-2019-semi}
Rui Cai and Mirella Lapata. 2019.
\newblock \href {https://doi.org/10.18653/v1/D19-1094} {Semi-supervised
  semantic role labeling with cross-view training}.
\newblock In \emph{Proceedings of the 2019 Conference on Empirical Methods in
  Natural Language Processing and the 9th International Joint Conference on
  Natural Language Processing (EMNLP-IJCNLP)}, pages 1018--1027, Hong Kong,
  China. Association for Computational Linguistics.

\bibitem[{Chang et~al.(2008)Chang, Ratinov, Roth, and
  Srikumar}]{chang2008importance}
Ming-Wei Chang, Lev Ratinov, Dan Roth, and Vivek Srikumar. 2008.
\newblock \href {https://www.aaai.org/Papers/AAAI/2008/AAAI08-132.pdf}
  {Importance of semantic representation: dataless classification}.
\newblock In \emph{Proceedings of the 23rd national conference on Artificial
  intelligence-Volume 2}, pages 830--835.

\bibitem[{Chapelle et~al.(2009)Chapelle, Scholkopf, and
  Zien}]{chapelle2009semi}
Olivier Chapelle, Bernhard Scholkopf, and Alexander Zien. 2009.
\newblock \href {https://ieeexplore.ieee.org/abstract/document/4787647}
  {Semi-supervised learning (chapelle, o. et al., eds.; 2006)[book reviews]}.
\newblock \emph{IEEE Transactions on Neural Networks}, 20(3):542--542.

\bibitem[{Chen et~al.(2022)Chen, Jiang, Wang, Wan, Wang, and Long}]{DST2022}
Baixu Chen, Junguang Jiang, Ximei Wang, Pengfei Wan, Jianmin Wang, and
  Mingsheng Long. 2022.
\newblock \href {https://openreview.net/forum?id=NI7moUOKtc} {Debiased
  self-training for semi-supervised learning}.
\newblock In \emph{Advances in Neural Information Processing Systems}, NIPS'22.

\bibitem[{Chen et~al.(2020{\natexlab{a}})Chen, Wang, Tian, Yang, and
  Yang}]{chen-etal-2020-local}
Jiaao Chen, Zhenghui Wang, Ran Tian, Zichao Yang, and Diyi Yang.
  2020{\natexlab{a}}.
\newblock \href {https://doi.org/10.18653/v1/2020.emnlp-main.95} {Local
  additivity based data augmentation for semi-supervised {NER}}.
\newblock In \emph{Proceedings of the 2020 Conference on Empirical Methods in
  Natural Language Processing (EMNLP)}, pages 1241--1251, Online. Association
  for Computational Linguistics.

\bibitem[{Chen et~al.(2020{\natexlab{b}})Chen, Yang, and
  Yang}]{chen-etal-2020-mixtext}
Jiaao Chen, Zichao Yang, and Diyi Yang. 2020{\natexlab{b}}.
\newblock \href {https://doi.org/10.18653/v1/2020.acl-main.194} {{M}ix{T}ext:
  Linguistically-informed interpolation of hidden space for semi-supervised
  text classification}.
\newblock In \emph{Proceedings of the 58th Annual Meeting of the Association
  for Computational Linguistics}, pages 2147--2157, Online. Association for
  Computational Linguistics.

\bibitem[{Clark et~al.(2018)Clark, Luong, Manning, and
  Le}]{clark-etal-2018-semi}
Kevin Clark, Minh-Thang Luong, Christopher~D. Manning, and Quoc Le. 2018.
\newblock \href {https://doi.org/10.18653/v1/D18-1217} {Semi-supervised
  sequence modeling with cross-view training}.
\newblock In \emph{Proceedings of the 2018 Conference on Empirical Methods in
  Natural Language Processing}, pages 1914--1925, Brussels, Belgium.
  Association for Computational Linguistics.

\bibitem[{Devlin et~al.(2019)Devlin, Chang, Lee, and
  Toutanova}]{devlin2018bert}
Jacob Devlin, Ming-Wei Chang, Kenton Lee, and Kristina Toutanova. 2019.
\newblock \href {https://doi.org/10.18653/v1/N19-1423} {{BERT}: Pre-training of
  deep bidirectional transformers for language understanding}.
\newblock In \emph{Proceedings of the 2019 Conference of the North {A}merican
  Chapter of the Association for Computational Linguistics: Human Language
  Technologies, Volume 1 (Long and Short Papers)}, Minneapolis, Minnesota.
  Association for Computational Linguistics.

\bibitem[{Dong and de~Melo(2019)}]{dong-de-melo-2019-robust}
Xin Dong and Gerard de~Melo. 2019.
\newblock \href {https://doi.org/10.18653/v1/D19-1658} {A robust self-learning
  framework for cross-lingual text classification}.
\newblock In \emph{Proceedings of the 2019 Conference on Empirical Methods in
  Natural Language Processing and the 9th International Joint Conference on
  Natural Language Processing (EMNLP-IJCNLP)}, pages 6306--6310, Hong Kong,
  China. Association for Computational Linguistics.

\bibitem[{Gao et~al.(2021)Gao, Fisch, and Chen}]{gao-etal-2021-making}
Tianyu Gao, Adam Fisch, and Danqi Chen. 2021.
\newblock \href {https://doi.org/10.18653/v1/2021.acl-long.295} {Making
  pre-trained language models better few-shot learners}.
\newblock In \emph{Proceedings of the 59th Annual Meeting of the Association
  for Computational Linguistics and the 11th International Joint Conference on
  Natural Language Processing (Volume 1: Long Papers)}, pages 3816--3830,
  Online. Association for Computational Linguistics.

\bibitem[{Gera et~al.(2022)Gera, Halfon, Shnarch, Perlitz, Ein-Dor, and
  Slonim}]{gera2022zero}
Ariel Gera, Alon Halfon, Eyal Shnarch, Yotam Perlitz, Liat Ein-Dor, and Noam
  Slonim. 2022.
\newblock \href
  {https://research.ibm.com/publications/zero-shot-text-classification-with-self-training}
  {Zero-shot text classification with self-training}.
\newblock In \emph{Proceedings of the 2022 Conference on Empirical Methods in
  Natural Language Processing}. Association for Computational Linguistics.

\bibitem[{Goel et~al.(2022)Goel, Jiao, and Massiah}]{goel2022pars}
Arushi Goel, Yunlong Jiao, and Jordan Massiah. 2022.
\newblock \href {https://arxiv.org/pdf/2201.10836.pdf} {Pars: Pseudo-label
  aware robust sample selection for learning with noisy labels}.
\newblock \emph{arXiv preprint arXiv:2201.10836}.

\bibitem[{Grandvalet and Bengio(2004)}]{grandvalet2004semi}
Yves Grandvalet and Yoshua Bengio. 2004.
\newblock \href
  {https://proceedings.neurips.cc/paper/2004/file/96f2b50b5d3613adf9c27049b2a888c7-Paper.pdf}
  {Semi-supervised learning by entropy minimization}.
\newblock \emph{Advances in neural information processing systems}, 17.

\bibitem[{Gururangan et~al.(2019)Gururangan, Dang, Card, and
  Smith}]{gururangan-etal-2019-variational}
Suchin Gururangan, Tam Dang, Dallas Card, and Noah~A. Smith. 2019.
\newblock \href {https://doi.org/10.18653/v1/P19-1590} {Variational pretraining
  for semi-supervised text classification}.
\newblock In \emph{Proceedings of the 57th Annual Meeting of the Association
  for Computational Linguistics}, pages 5880--5894, Florence, Italy.
  Association for Computational Linguistics.

\bibitem[{Gururangan et~al.(2020)Gururangan, Marasovi{\'c}, Swayamdipta, Lo,
  Beltagy, Downey, and Smith}]{gururangan-etal-2020-dont}
Suchin Gururangan, Ana Marasovi{\'c}, Swabha Swayamdipta, Kyle Lo, Iz~Beltagy,
  Doug Downey, and Noah~A. Smith. 2020.
\newblock \href {https://doi.org/10.18653/v1/2020.acl-main.740} {Don{'}t stop
  pretraining: Adapt language models to domains and tasks}.
\newblock In \emph{Proceedings of the 58th Annual Meeting of the Association
  for Computational Linguistics}, pages 8342--8360, Online. Association for
  Computational Linguistics.

\bibitem[{He et~al.(2020)He, Gu, Shen, and Ranzato}]{He2020Revisiting}
Junxian He, Jiatao Gu, Jiajun Shen, and Marc'Aurelio Ranzato. 2020.
\newblock \href {https://openreview.net/forum?id=SJgdnAVKDH} {Revisiting
  self-training for neural sequence generation}.
\newblock In \emph{International Conference on Learning Representations}.

\bibitem[{Hendriksen et~al.(2022)Hendriksen, Bleeker, Vakulenko, van Noord,
  Kuiper, and de~Rijke}]{10.1007/978-3-030-99736-6_20}
Mariya Hendriksen, Maurits Bleeker, Svitlana Vakulenko, Nanne van Noord, Ernst
  Kuiper, and Maarten de~Rijke. 2022.
\newblock \href {https://doi.org/10.1007/978-3-030-99736-6_20} {Extending clip
  for category-to-image retrieval in e-commerce}.
\newblock In \emph{Advances in Information Retrieval: 44th European Conference
  on IR Research, ECIR 2022, Stavanger, Norway, April 10–14, 2022,
  Proceedings, Part I}, page 289–303, Berlin, Heidelberg. Springer-Verlag.

\bibitem[{Hoffmann et~al.(2022)Hoffmann, Borgeaud, Mensch, Buchatskaya, Cai,
  Rutherford, Casas, Hendricks, Welbl, Clark et~al.}]{hoffmann2022training}
Jordan Hoffmann, Sebastian Borgeaud, Arthur Mensch, Elena Buchatskaya, Trevor
  Cai, Eliza Rutherford, Diego de~Las Casas, Lisa~Anne Hendricks, Johannes
  Welbl, Aidan Clark, et~al. 2022.
\newblock \href
  {https://www.deepmind.com/publications/an-empirical-analysis-of-compute-optimal-large-language-model-training}
  {Training compute-optimal large language models}.
\newblock \emph{arXiv preprint arXiv:2203.15556}.

\bibitem[{Hou et~al.(2022)Hou, Salazar, and Polovets}]{10.1162/tacl_a_00517}
Zejiang Hou, Julian Salazar, and George Polovets. 2022.
\newblock \href {https://doi.org/10.1162/tacl_a_00517} {{Meta-Learning the
  Difference: Preparing Large Language Models for Efficient Adaptation}}.
\newblock \emph{Transactions of the Association for Computational Linguistics},
  10:1249--1265.

\bibitem[{Howard and Ruder(2018)}]{howard-ruder-2018-universal}
Jeremy Howard and Sebastian Ruder. 2018.
\newblock \href {https://doi.org/10.18653/v1/P18-1031} {Universal language
  model fine-tuning for text classification}.
\newblock In \emph{Proceedings of the 56th Annual Meeting of the Association
  for Computational Linguistics (Volume 1: Long Papers)}, pages 328--339,
  Melbourne, Australia. Association for Computational Linguistics.

\bibitem[{Kaplan et~al.(2020)Kaplan, McCandlish, Henighan, Brown, Chess, Child,
  Gray, Radford, Wu, and Amodei}]{kaplan2020scaling}
Jared Kaplan, Sam McCandlish, Tom Henighan, Tom~B Brown, Benjamin Chess, Rewon
  Child, Scott Gray, Alec Radford, Jeffrey Wu, and Dario Amodei. 2020.
\newblock \href {https://arxiv.org/abs/2001.08361} {Scaling laws for neural
  language models}.
\newblock \emph{arXiv preprint arXiv:2001.08361}.

\bibitem[{Kedzie and McKeown(2019)}]{kedzie-mckeown-2019-good}
Chris Kedzie and Kathleen McKeown. 2019.
\newblock \href {https://doi.org/10.18653/v1/W19-8672} {A good sample is hard
  to find: Noise injection sampling and self-training for neural language
  generation models}.
\newblock In \emph{Proceedings of the 12th International Conference on Natural
  Language Generation}, pages 584--593, Tokyo, Japan. Association for
  Computational Linguistics.

\bibitem[{Kipf and Welling(2017)}]{kipf2017semi}
Thomas~N. Kipf and Max Welling. 2017.
\newblock \href {https://openreview.net/pdf?id=SJU4ayYgl} {Semi-supervised
  classification with graph convolutional networks}.
\newblock In \emph{International Conference on Learning Representations
  (ICLR)}.

\bibitem[{Laine and Aila(2017)}]{DBLP:conf/iclr/LaineA17}
Samuli Laine and Timo Aila. 2017.
\newblock \href {https://openreview.net/forum?id=BJ6oOfqge} {Temporal
  ensembling for semi-supervised learning}.
\newblock In \emph{5th International Conference on Learning Representations,
  {ICLR} 2017, Toulon, France, April 24-26, 2017, Conference Track
  Proceedings}. OpenReview.net.

\bibitem[{Lee et~al.(2013)}]{lee2013pseudo}
Dong-Hyun Lee et~al. 2013.
\newblock \href
  {https://www.kaggle.com/blobs/download/forum-message-attachment-files/746/pseudo_label_final.pdf}
  {Pseudo-label: The simple and efficient semi-supervised learning method for
  deep neural networks}.
\newblock In \emph{Workshop on challenges in representation learning, ICML},
  page 896.

\bibitem[{Li et~al.(2021{\natexlab{a}})Li, Li, and
  Ouyang}]{li-etal-2021-semi-supervised}
Changchun Li, Ximing Li, and Jihong Ouyang. 2021{\natexlab{a}}.
\newblock \href {https://doi.org/10.18653/v1/2021.acl-long.391}
  {Semi-supervised text classification with balanced deep representation
  distributions}.
\newblock In \emph{Proceedings of the 59th Annual Meeting of the Association
  for Computational Linguistics and the 11th International Joint Conference on
  Natural Language Processing (Volume 1: Long Papers)}, pages 5044--5053,
  Online. Association for Computational Linguistics.

\bibitem[{Li et~al.(2021{\natexlab{b}})Li, Yavuz, Chen, and
  Yan}]{li-etal-2021-task-adaptive}
Shiyang Li, Semih Yavuz, Wenhu Chen, and Xifeng Yan. 2021{\natexlab{b}}.
\newblock \href {https://doi.org/10.18653/v1/2021.findings-emnlp.86}
  {Task-adaptive pre-training and self-training are complementary for natural
  language understanding}.
\newblock In \emph{Findings of the Association for Computational Linguistics:
  EMNLP 2021}, pages 1006--1015, Punta Cana, Dominican Republic. Association
  for Computational Linguistics.

\bibitem[{Li et~al.(2011)Li, Wang, Zhou, and Lee}]{10.5555/2283696.2283708}
Shoushan Li, Zhongqing Wang, Guodong Zhou, and Sophia Yat~Mei Lee. 2011.
\newblock \href {https://www.ijcai.org/Proceedings/11/Papers/306.pdf}
  {Semi-supervised learning for imbalanced sentiment classification}.
\newblock In \emph{Proceedings of the Twenty-Second International Joint
  Conference on Artificial Intelligence - Volume Volume Three}, IJCAI'11, page
  1826–1831. AAAI Press.

\bibitem[{Li et~al.(2019)Li, Peng, Zhang, Wang, and
  Si}]{li_semi-supervised_2019}
Zhenghua Li, Xue Peng, Min Zhang, Rui Wang, and Luo Si. 2019.
\newblock \href {https://doi.org/10.18653/v1/P19-1229} {Semi-supervised
  {Domain} {Adaptation} for {Dependency} {Parsing}}.
\newblock In \emph{Proceedings of the 57th {Annual} {Meeting} of the
  {Association} for {Computational} {Linguistics}}, pages 2386--2395, Florence,
  Italy. Association for Computational Linguistics.

\bibitem[{Liu et~al.(2019)Liu, Ott, Goyal, Du, Joshi, Chen, Levy, Lewis,
  Zettlemoyer, and Stoyanov}]{liu2019roberta}
Yinhan Liu, Myle Ott, Naman Goyal, Jingfei Du, Mandar Joshi, Danqi Chen, Omer
  Levy, Mike Lewis, Luke Zettlemoyer, and Veselin Stoyanov. 2019.
\newblock \href {https://arxiv.org/abs/1907.11692} {Roberta: A robustly
  optimized bert pretraining approach}.
\newblock \emph{arXiv preprint arXiv:1907.11692}.

\bibitem[{Logeswaran et~al.(2019)Logeswaran, Chang, Lee, Toutanova, Devlin, and
  Lee}]{logeswaran-etal-2019-zero}
Lajanugen Logeswaran, Ming-Wei Chang, Kenton Lee, Kristina Toutanova, Jacob
  Devlin, and Honglak Lee. 2019.
\newblock \href {https://doi.org/10.18653/v1/P19-1335} {Zero-shot entity
  linking by reading entity descriptions}.
\newblock In \emph{Proceedings of the 57th Annual Meeting of the Association
  for Computational Linguistics}, pages 3449--3460, Florence, Italy.
  Association for Computational Linguistics.

\bibitem[{Maas et~al.(2011)Maas, Daly, Pham, Huang, Ng, and
  Potts}]{maas-etal-2011-learning}
Andrew~L. Maas, Raymond~E. Daly, Peter~T. Pham, Dan Huang, Andrew~Y. Ng, and
  Christopher Potts. 2011.
\newblock \href {https://aclanthology.org/P11-1015} {Learning word vectors for
  sentiment analysis}.
\newblock In \emph{Proceedings of the 49th Annual Meeting of the Association
  for Computational Linguistics: Human Language Technologies}, pages 142--150,
  Portland, USA. Association for Computational Linguistics.

\bibitem[{Margatina et~al.(2022)Margatina, Barrault, and
  Aletras}]{margatina-etal-2022-importance}
Katerina Margatina, Loic Barrault, and Nikolaos Aletras. 2022.
\newblock \href {https://doi.org/10.18653/v1/2022.acl-short.93} {On the
  importance of effectively adapting pretrained language models for active
  learning}.
\newblock In \emph{Proceedings of the 60th Annual Meeting of the Association
  for Computational Linguistics (Volume 2: Short Papers)}, pages 825--836,
  Dublin, Ireland. Association for Computational Linguistics.

\bibitem[{McAuley and Leskovec(2013)}]{10.1145/2507157.2507163}
Julian McAuley and Jure Leskovec. 2013.
\newblock \href {https://doi.org/10.1145/2507157.2507163} {Hidden factors and
  hidden topics: Understanding rating dimensions with review text}.
\newblock In \emph{Proceedings of the 7th ACM Conference on Recommender
  Systems}, RecSys '13, page 165–172, New York, NY, USA. Association for
  Computing Machinery.

\bibitem[{McClosky et~al.(2006)McClosky, Charniak, and
  Johnson}]{mcclosky-etal-2006-effective}
David McClosky, Eugene Charniak, and Mark Johnson. 2006.
\newblock \href {https://aclanthology.org/N06-1020} {Effective self-training
  for parsing}.
\newblock In \emph{Proceedings of the Human Language Technology Conference of
  the {NAACL}, Main Conference}, pages 152--159, New York City, USA.
  Association for Computational Linguistics.

\bibitem[{Mehri et~al.(2020)Mehri, Eric, and
  Hakkani-T{\"u}r}]{Mehri2020DialoGLUEAN}
Shikib Mehri, Mihail Eric, and Dilek~Z. Hakkani-T{\"u}r. 2020.
\newblock \href {https://arxiv.org/pdf/2009.13570.pdf} {Dialoglue: A natural
  language understanding benchmark for task-oriented dialogue}.
\newblock \emph{ArXiv}, abs/2009.13570.

\bibitem[{Meng et~al.(2020)Meng, Zhang, Huang, Xiong, Ji, Zhang, and
  Han}]{meng-etal-2020-text}
Yu~Meng, Yunyi Zhang, Jiaxin Huang, Chenyan Xiong, Heng Ji, Chao Zhang, and
  Jiawei Han. 2020.
\newblock \href {https://doi.org/10.18653/v1/2020.emnlp-main.724} {Text
  classification using label names only: A language model self-training
  approach}.
\newblock In \emph{Proceedings of the 2020 Conference on Empirical Methods in
  Natural Language Processing (EMNLP)}, pages 9006--9017, Online. Association
  for Computational Linguistics.

\bibitem[{Miyato et~al.(2018)Miyato, Maeda, Koyama, and
  Ishii}]{miyato2018virtual}
Takeru Miyato, Shin-ichi Maeda, Masanori Koyama, and Shin Ishii. 2018.
\newblock \href {https://arxiv.org/pdf/1704.03976.pdf} {Virtual adversarial
  training: a regularization method for supervised and semi-supervised
  learning}.
\newblock \emph{IEEE transactions on pattern analysis and machine
  intelligence}, 41:1979--1993.

\bibitem[{Mukherjee and Awadallah(2020)}]{NEURIPS2020_f23d125d}
Subhabrata Mukherjee and Ahmed Awadallah. 2020.
\newblock \href
  {https://proceedings.neurips.cc/paper/2020/file/f23d125da1e29e34c552f448610ff25f-Paper.pdf}
  {Uncertainty-aware self-training for few-shot text classification}.
\newblock In \emph{Advances in Neural Information Processing Systems},
  volume~33, pages 21199--21212. Curran Associates, Inc.

\bibitem[{Ott et~al.(2019)Ott, Edunov, Baevski, Fan, Gross, Ng, Grangier, and
  Auli}]{ott2019fairseq}
Myle Ott, Sergey Edunov, Alexei Baevski, Angela Fan, Sam Gross, Nathan Ng,
  David Grangier, and Michael Auli. 2019.
\newblock \href {https://github.com/facebookresearch/fairseq} {fairseq: A fast,
  extensible toolkit for sequence modeling}.
\newblock In \emph{Proceedings of NAACL-HLT 2019: Demonstrations}.

\bibitem[{Radford et~al.(2019)Radford, Wu, Child, Luan, Amodei, Sutskever
  et~al.}]{radford2019language}
Alec Radford, Jeffrey Wu, Rewon Child, David Luan, Dario Amodei, Ilya
  Sutskever, et~al. 2019.
\newblock \href
  {https://cdn.openai.com/better-language-models/language_models_are_unsupervised_multitask_learners.pdf}
  {Language models are unsupervised multitask learners}.
\newblock \emph{OpenAI blog}, 1(8).

\bibitem[{Raina et~al.(2007)Raina, Battle, Lee, Packer, and Ng}]{raina2007self}
Rajat Raina, Alexis Battle, Honglak Lee, Benjamin Packer, and Andrew~Y. Ng.
  2007.
\newblock \href {https://doi.org/10.1145/1273496.1273592} {Self-taught
  learning: Transfer learning from unlabeled data}.
\newblock In \emph{Proceedings of the 24th International Conference on Machine
  Learning}, ICML '07, page 759–766, New York, NY, USA. Association for
  Computing Machinery.

\bibitem[{Ramponi and Plank(2020)}]{ramponi_neural_2020}
Alan Ramponi and Barbara Plank. 2020.
\newblock \href {https://doi.org/10.18653/v1/2020.coling-main.603} {Neural
  {Unsupervised} {Domain} {Adaptation} in {NLP}—{A} {Survey}}.
\newblock In \emph{Proceedings of the 28th {International} {Conference} on
  {Computational} {Linguistics}}, pages 6838--6855, Barcelona, Spain (Online).
  International Committee on Computational Linguistics.

\bibitem[{Ruder and Plank(2018)}]{ruder_strong_2018}
Sebastian Ruder and Barbara Plank. 2018.
\newblock \href {https://doi.org/10.18653/v1/P18-1096} {Strong {Baselines} for
  {Neural} {Semi}-{Supervised} {Learning} under {Domain} {Shift}}.
\newblock In \emph{Proceedings of the 56th {Annual} {Meeting} of the
  {Association} for {Computational} {Linguistics} ({Volume} 1: {Long}
  {Papers})}, pages 1044--1054, Melbourne, Australia. Association for
  Computational Linguistics.

\bibitem[{Saito et~al.(2018)Saito, Watanabe, Ushiku, and
  Harada}]{saito2018maximum}
Kuniaki Saito, Kohei Watanabe, Yoshitaka Ushiku, and Tatsuya Harada. 2018.
\newblock \href
  {https://openaccess.thecvf.com/content_cvpr_2018/html/Saito_Maximum_Classifier_Discrepancy_CVPR_2018_paper.html}
  {Maximum classifier discrepancy for unsupervised domain adaptation}.
\newblock In \emph{Proceedings of the IEEE conference on computer vision and
  pattern recognition}, pages 3723--3732.

\bibitem[{Schick and Sch{\"u}tze(2021)}]{schick-schutze-2021-exploiting}
Timo Schick and Hinrich Sch{\"u}tze. 2021.
\newblock \href {https://doi.org/10.18653/v1/2021.eacl-main.20} {Exploiting
  cloze-questions for few-shot text classification and natural language
  inference}.
\newblock In \emph{Proceedings of the 16th Conference of the European Chapter
  of the Association for Computational Linguistics: Main Volume}, pages
  255--269, Online. Association for Computational Linguistics.

\bibitem[{Shi et~al.(2022{\natexlab{a}})Shi, Feng, and
  Lipani}]{shi-etal-2022-learning}
Zhengxiang Shi, Yue Feng, and Aldo Lipani. 2022{\natexlab{a}}.
\newblock \href {https://doi.org/10.18653/v1/2022.findings-naacl.158} {Learning
  to execute actions or ask clarification questions}.
\newblock In \emph{Findings of the Association for Computational Linguistics:
  NAACL 2022}, pages 2060--2070, Seattle, United States. Association for
  Computational Linguistics.

\bibitem[{Shi and Lipani(2023)}]{shi2023don}
Zhengxiang Shi and Aldo Lipani. 2023.
\newblock \href {https://arxiv.org/pdf/2305.01711.pdf} {Don't stop pretraining?
  make prompt-based fine-tuning powerful learner}.
\newblock \emph{arXiv preprint arXiv:2305.01711}.

\bibitem[{Shi et~al.(2022{\natexlab{b}})Shi, Zhang, and
  Lipani}]{stepGame2022shi}
Zhengxiang Shi, Qiang Zhang, and Aldo Lipani. 2022{\natexlab{b}}.
\newblock \href {https://doi.org/10.1609/aaai.v36i10.21383} {Stepgame: A new
  benchmark for robust multi-hop spatial reasoning in texts}.
\newblock In \emph{Proceedings of the AAAI Conference on Artificial
  Intelligence}, volume~36, pages 11321--11329.

\bibitem[{Sohn et~al.(2020)Sohn, Berthelot, Li, Zhang, Carlini, Cubuk, Kurakin,
  Zhang, and Raffel}]{sohn2020fixmatch}
Kihyuk Sohn, David Berthelot, Chun-Liang Li, Zizhao Zhang, Nicholas Carlini,
  Ekin~D. Cubuk, Alex Kurakin, Han Zhang, and Colin Raffel. 2020.
\newblock \href
  {https://papers.nips.cc/paper/2020/file/06964dce9addb1c5cb5d6e3d9838f733-Paper.pdf}
  {Fixmatch: Simplifying semi-supervised learning with consistency and
  confidence}.
\newblock In \emph{Proceedings of the 34th International Conference on Neural
  Information Processing Systems}, NIPS'20, Red Hook, NY, USA. Curran
  Associates Inc.

\bibitem[{Tarvainen and Valpola(2017)}]{tarvainen2017mean}
Antti Tarvainen and Harri Valpola. 2017.
\newblock \href
  {https://proceedings.neurips.cc/paper/2017/file/68053af2923e00204c3ca7c6a3150cf7-Paper.pdf}
  {Mean teachers are better role models: Weight-averaged consistency targets
  improve semi-supervised deep learning results}.
\newblock In \emph{Proceedings of the 31st International Conference on Neural
  Information Processing Systems}, NIPS'17, page 1195–1204, Red Hook, NY,
  USA. Curran Associates Inc.

\bibitem[{Wang et~al.(2018)Wang, Singh, Michael, Hill, Levy, and
  Bowman}]{wang-etal-2018-glue}
Alex Wang, Amanpreet Singh, Julian Michael, Felix Hill, Omer Levy, and Samuel
  Bowman. 2018.
\newblock \href {https://doi.org/10.18653/v1/W18-5446} {{GLUE}: A multi-task
  benchmark and analysis platform for natural language understanding}.
\newblock In \emph{Proceedings of the 2018 {EMNLP} Workshop {B}lackbox{NLP}:
  Analyzing and Interpreting Neural Networks for {NLP}}, pages 353--355,
  Brussels, Belgium. Association for Computational Linguistics.

\bibitem[{Wang et~al.(2021)Wang, Gao, Long, and Wang}]{wang2021selftuning}
Ximei Wang, Jinghan Gao, Mingsheng Long, and Jianmin Wang. 2021.
\newblock \href {https://arxiv.org/abs/2102.12903} {Self-tuning for
  data-efficient deep learning}.
\newblock In \emph{International Conference on Machine Learning (ICML)}.

\bibitem[{Wang et~al.(2022)Wang, Chen, Fan, SUN, Tao, Hou, Wang, Yang, Zhou,
  Guo, Qi, Wu, Li, Nakamura, Ye, Savvides, Raj, Shinozaki, Schiele, Wang, Xie,
  and Zhang}]{wang2022usb}
Yidong Wang, Hao Chen, Yue Fan, Wang SUN, Ran Tao, Wenxin Hou, Renjie Wang,
  Linyi Yang, Zhi Zhou, Lan-Zhe Guo, Heli Qi, Zhen Wu, Yu-Feng Li, Satoshi
  Nakamura, Wei Ye, Marios Savvides, Bhiksha Raj, Takahiro Shinozaki, Bernt
  Schiele, Jindong Wang, Xing Xie, and Yue Zhang. 2022.
\newblock \href {https://openreview.net/forum?id=QeuwINa96C} {{USB}: A unified
  semi-supervised learning benchmark for classification}.
\newblock In \emph{Thirty-sixth Conference on Neural Information Processing
  Systems Datasets and Benchmarks Track}.

\bibitem[{Xie et~al.(2020{\natexlab{a}})Xie, Dai, Hovy, Luong, and
  Le}]{10.5555/3495724.3496249}
Qizhe Xie, Zihang Dai, Eduard Hovy, Minh-Thang Luong, and Quoc~V. Le.
  2020{\natexlab{a}}.
\newblock \href
  {https://proceedings.neurips.cc/paper/2020/file/44feb0096faa8326192570788b38c1d1-Paper.pdf}
  {Unsupervised data augmentation for consistency training}.
\newblock In \emph{Proceedings of the 34th International Conference on Neural
  Information Processing Systems}, NIPS'20, Red Hook, NY, USA. Curran
  Associates Inc.

\bibitem[{Xie et~al.(2020{\natexlab{b}})Xie, Luong, Hovy, and Le}]{xie2020self}
Qizhe Xie, Minh-Thang Luong, Eduard Hovy, and Quoc~V Le. 2020{\natexlab{b}}.
\newblock \href
  {https://openaccess.thecvf.com/content_CVPR_2020/papers/Xie_Self-Training_With_Noisy_Student_Improves_ImageNet_Classification_CVPR_2020_paper.pdf}
  {Self-training with noisy student improves imagenet classification}.
\newblock In \emph{Proceedings of the IEEE/CVF conference on computer vision
  and pattern recognition}, pages 10687--10698.

\bibitem[{Xu et~al.(2021{\natexlab{a}})Xu, Baevski, Likhomanenko, Tomasello,
  Conneau, Collobert, Synnaeve, and Auli}]{xu2021self}
Qiantong Xu, Alexei Baevski, Tatiana Likhomanenko, Paden Tomasello, Alexis
  Conneau, Ronan Collobert, Gabriel Synnaeve, and Michael Auli.
  2021{\natexlab{a}}.
\newblock \href {https://ieeexplore.ieee.org/stamp/stamp.jsp?arnumber=9414641}
  {Self-training and pre-training are complementary for speech recognition}.
\newblock In \emph{ICASSP 2021-2021 IEEE International Conference on Acoustics,
  Speech and Signal Processing (ICASSP)}, pages 3030--3034. IEEE.

\bibitem[{Xu et~al.(2021{\natexlab{b}})Xu, Shang, Ye, Qian, Li, Sun, Li, and
  Jin}]{xu2021dash}
Yi~Xu, Lei Shang, Jinxing Ye, Qi~Qian, Yu-Feng Li, Baigui Sun, Hao Li, and Rong
  Jin. 2021{\natexlab{b}}.
\newblock \href {http://proceedings.mlr.press/v139/xu21e/xu21e.pdf} {Dash:
  Semi-supervised learning with dynamic thresholding}.
\newblock In \emph{International Conference on Machine Learning}, pages
  11525--11536. PMLR.

\bibitem[{Xue et~al.(2021)Xue, Constant, Roberts, Kale, Al-Rfou, Siddhant,
  Barua, and Raffel}]{xue-etal-2021-mt5}
Linting Xue, Noah Constant, Adam Roberts, Mihir Kale, Rami Al-Rfou, Aditya
  Siddhant, Aditya Barua, and Colin Raffel. 2021.
\newblock \href {https://doi.org/10.18653/v1/2021.naacl-main.41} {m{T}5: A
  massively multilingual pre-trained text-to-text transformer}.
\newblock In \emph{Proceedings of the 2021 Conference of the North American
  Chapter of the Association for Computational Linguistics: Human Language
  Technologies}, pages 483--498, Online. Association for Computational
  Linguistics.

\bibitem[{Yarowsky(1995)}]{yarowsky-1995-unsupervised}
David Yarowsky. 1995.
\newblock \href {https://doi.org/10.3115/981658.981684} {Unsupervised word
  sense disambiguation rivaling supervised methods}.
\newblock In \emph{33rd Annual Meeting of the Association for Computational
  Linguistics}, pages 189--196, Cambridge, Massachusetts, USA. Association for
  Computational Linguistics.

\bibitem[{Zhang et~al.(2021)Zhang, Wang, Hou, Wu, Wang, Okumura, and
  Shinozaki}]{zhang2021flexmatch}
Bowen Zhang, Yidong Wang, Wenxin Hou, Hao Wu, Jindong Wang, Manabu Okumura, and
  Takahiro Shinozaki. 2021.
\newblock \href
  {https://proceedings.neurips.cc/paper/2021/hash/995693c15f439e3d189b06e89d145dd5-Abstract.html}
  {Flexmatch: Boosting semi-supervised learning with curriculum pseudo
  labeling}.
\newblock In \emph{Proceedings of the 35th International Conference on Neural
  Information Processing Systems}, volume~34.

\bibitem[{Zhang et~al.(2015)Zhang, Zhao, and LeCun}]{10.5555/2969239.2969312}
Xiang Zhang, Junbo Zhao, and Yann LeCun. 2015.
\newblock \href
  {https://proceedings.neurips.cc/paper/2015/file/250cf8b51c773f3f8dc8b4be867a9a02-Paper.pdf}
  {Character-level convolutional networks for text classification}.
\newblock In \emph{Proceedings of the 28th International Conference on Neural
  Information Processing Systems - Volume 1}, NIPS'15, page 649–657,
  Cambridge, MA, USA. MIT Press.

\bibitem[{Zoph et~al.(2020)Zoph, Ghiasi, Lin, Cui, Liu, Cubuk, and
  Le}]{10.5555/3495724.3496047}
Barret Zoph, Golnaz Ghiasi, Tsung-Yi Lin, Yin Cui, Hanxiao Liu, Ekin~D. Cubuk,
  and Quoc~V. Le. 2020.
\newblock \href
  {https://proceedings.neurips.cc/paper/2020/file/27e9661e033a73a6ad8cefcde965c54d-Paper.pdf}
  {Rethinking pre-training and self-training}.
\newblock In \emph{Proceedings of the 34th International Conference on Neural
  Information Processing Systems}, NIPS'20, Red Hook, NY, USA. Curran
  Associates Inc.

\end{thebibliography}
